\documentclass[12pt,onecolumn]{IEEEtran}

\usepackage{floatflt}
\usepackage{multirow}
\usepackage{hyperref}
\usepackage{graphicx}
\usepackage{url}
\usepackage{color}
\usepackage{subfigure}
\usepackage{amssymb}
\usepackage{amsmath}
\usepackage{times}
\usepackage[pagewise]{lineno}

\def\a{\mathbf a}

\def\c{\mathbf c}

\def\e{\mathbf e}

\def\p{\mathbf p}

\def\v{\mathbf v}
\def\w{\mathbf w}
\def\x{\mathbf x}
\def\y{\mathbf y}
\def\z{\mathbf z}

\newtheorem{definition}{Definition}

\newtheorem{theorem}{Theorem}
\DeclareMathOperator*{\argmax}{argmax}

\title{Post-Selections in AI and How to Avoid Them}

\author{
\IEEEauthorblockN{Juyang Weng\IEEEauthorrefmark{1}\IEEEauthorrefmark{2}\IEEEauthorrefmark{3}\IEEEauthorrefmark{4}\\
\IEEEauthorblockA{\IEEEauthorrefmark{1}Department of Computer Science and Engineering,\\ 
\IEEEauthorrefmark{2}Cognitive Science Program,\\
\IEEEauthorrefmark{3} Neuroscience Program, \\
Michigan State University, East Lansing, MI, 48824 USA}\\
\IEEEauthorrefmark{4} GENISAMA LLC, 4460 Alderwood Drive, Okemos, Michigan 48864 USA}
}

\begin{document}
\maketitle



{\bf Abstract:}  Neural network based Artificial Intelligence (AI) has reported increasing scales in experiments.  However, this paper raises a rarely reported stage in such experiments called Post-Selection alter the reader to 
several possible protocol flaws that may result in misleading results.   All AI methods fall into two broad
schools, connectionist and symbolic.  The Post-Selection fall into two kinds, Post-Selection Using Validation Sets (PSUVS) and Post-Selection Using Test Sets (PSUTS).  Each kind has two types of post-selectors, machines and humans.  The connectionist school received criticisms for its ``black box'' and now the Post-Selection; but the seemingly ``clean'' symbolic school seems more brittle because of its human PSUTS.   This paper first presents a controversial view: all static ``big data'' are non-scalable.  We then analyze why error-backprop from randomly initialized weights suffers from severe local minima, why PSUVS lacks cross-validation, why PSUTS violates well-established protocols, and why every paper involved should transparently report the Post-Selection stage.  To avoid future pitfalls in AI competitions, this paper proposes a new AI metrics, called developmental errors for all networks trained, under Three Learning Conditions: (1) an incremental learning architecture (due to a ``big data'' flaw), (2) a training experience and (3) a limited amount of computational resources.  Developmental Networks avoid Post-Selections because they automatically discover context-rules on the fly by generating emergent Turing machines (not black boxes) that are optimal in the sense of maximum-likelihood across lifetime, conditioned on the Three Learning Conditions.

%
%



%
%
%
\newpage
\setcounter{page}{2}

\section{Introduction}
\label{SE:intro}
AI research dates back at least to early 1910 when Leonardo Torres y Quevedo built a chess end game player 
called EI Ajedrecista \cite{Montfort05}.   In 1950, Alan Turing published his now celebrated paper \cite{Turing50} titled {\em Computing Machinery and Intelligence}.   Turing \cite{Turing50} was impressive to have discussed 
a wide variety of considerations for machine intelligence, as many as nine categories.  Unfortunately, he suggested to consider what is now called the Turing Test that has inspired and misled many AI researchers.

Much progress has been made in AI since then and many methods have been developed to deal with AI problems.   As the scope of this paper, we will focus on generalization. All AI methods fall into two schools \cite{WengRepRev12}, symbolic and connectionist, although many published methods are a mixture of both. 

 \subsection{Symbolic school}
 \label{SE:symbolicSchool}

Symbols are used in many AI methods (e.g., states in HMMs, nodes in Graphical Models and attributes in SLAM).  Although symbols are intuitive to a human programmer since he defines the associated meanings, symbols are static and have some fundamental limitations that have not received sufficient attention.   

The symbolic school \cite{Russell10} assumes a micro-world in 4D space-time in which a set of objects or concepts, e.g., $L=\{l_1, l_2, ... , l_n\}$, is assumed to be uniquely defined among 
many human programmers and their computers, represented by a series of symbols in time $\{ l_1(t), l_2(t), ... l_n(t)\;|\; t_0 \le t < t_1\}$.    The correspondences among all these symbols $\{l_i\}$ of the same object across different times are known as ``the frame problem'' \cite{Russell10} in AI which means that the programmer must manually link every symbol along time with its corresponding physical object.  In computer vision, the symbolic school assumes a single symbol  $o_i$, for all its 3D positions in its 3D trajectory $\{\x(t) \;|\; t_0\le t \le t_1\}$ and uses certain techniques, such as feature tracking through video (e.g., for driverless cars). Therefore, the symbolic school is based on human-handcrafted set of symbols and their assumed meanings.   Marvin Minsky wrote that symbols are ``neat'' \cite{Minsky91}, but in fact, symbols are ``neat'' mainly in a single human programmer's understanding but not between different programmers and not in relating computer programs to a real world.  

We will see the Developmental Network (DN) model of a brain is free from any symbols in its full version.  Abstract symbols correspond to action/state vectors in the motor area of DN.  Therefore, the frame problem is automatically solved through emergent action/state vectors in a physically grounded DN, without using any symbols in the DN's internal representations.

A major problem for symbolic AI is the generalization issue of symbols as defined here.

\begin{definition}[Brittleness of static symbols]
Suppose a symbolic AI machine $M(L)$ designed for a handcrafted set $L$ of symbols is applied to a real world that requires a new set of symbols $L'$, with $L\cap L'\ne \emptyset$,
$M(L)$ fails without a human programmer who handcrafts an appropriate mapping function $f: L' \mapsto L$ that maps every element of $L'$ to an element in $L$ so that $M(f(L'))$ works correctly as before. 
\end{definition}

Many expert systems (e.g.,  CYC, WordNet and EDR \cite{Lenat95b}) and ``big data'' projects \cite{Gome14} require a human programmer to be in the loop of handcrafting such a mapping $f$ during deployment.  For example, an machine $M$ developed in Florida is deployed in Michigan but Michigan has snow but Florida does not.  Because it is extremely challenging for a human programmer to understand many implicit limitations of $M(L)$, the mapping $f$ that the human handcrafts typically makes $M(F(L'))$ fail, resulting in the well-known high brittleness of symbolic systems.  

Due to emergent representations as numeric vectors, a DN robot discussed below learns snow settings and the snow concept when it sees snow scenes for the first time, because there are no symbols that correspond to snow in DN's representations.  

In general, the developmental methods to be discussed below automatically address such new concept problems without a need for a human programmer to be in the loop of handcrafting a symbolic mapping $f$
during a deployment.   In this paper, the author will further argue that the symbolic school suffers from human PSUTS.

\subsection{Connectionist School}
\label{SE:connectionistSchool}

The connectionist school claimed to be less brittle \cite{Rumelhart86,McClelland86}.  However, a network is egocentric---meaning that the agent starts from its own (neural) network, instead of
a symbolic world.   It must learn from the external {\em world} without a handcrafted, {\em world-centered} object model.   Although connectionist methods often assume some task-specific symbols,
e.g., a static set $L$ of object labels, they also assume a restricted world implicitly.  Therefore, a connectionist model typically needs to sense and learn from a restricted world using a network.   The use of $L$ by any neural networks (e.g., ImageNet \cite{Frishna17} and many other competitions) as a set of object labels is a fundamental limitation that also causes the resulting system to be brittle for the same reason as the symbolic school.  

Typically, a neural network is meant to establish only a mapping $f$ from the space of input $X$ to the space of
class labels $L$, 
\begin{equation}
f:X \mapsto L
\label{EQ:XtoLmapping}
\end{equation}
\cite{Funahashi89,Poggio90a}.   $X$ many contains a few time frames.  Many video analysis problems, speech recognition problems, and computer game-play problems are also converted into this static input space so that the input space also includes $L$, so as to learn 
\begin{equation}
f:X \times L \mapsto L.
\label{EQ:XLtoLmapping}
\end{equation}

Without a pressure of performance characterization during learning other than the performance of the final network, a self-organization map (e.g., SOM) has been used often as an unsupervised but slow learning method \cite{Kohonen01,Fukushima80,Oja03} .

In contrast, with a pressure of performance characterization during learning, Cresceptron~\cite{WengCresIJCV97} used a ``skull-closed'' incremental-learning Hebbian-like scheme with receptive-field based competitions. 

Other than the Hebbian mechanisms which are strictly ``unsupervised'' used by he Cresceptron and the DN explained below, two other types of learning schemes have been published:
\begin{description}

\item[A] Human handpicking features: after knowing the test set, humans handpick features, reported explicitly \cite{Fukushima83,Serre06,Fei-Fei06} or implicitly as ``weakly
supervised'' \cite{Fei-Fei06}.  This author called them ``skull-open'' \cite{WengAMDNL-2-2012}. 

\item[B] Error-backprop: Locally train multiple networks each from a different set of random weights. After the training, post-select the luckiest network.   Report the luckiest network only   \cite{Werbos94,LeCun98,Krizhevsky12,LeCun15,Mnih15,Silver16,Graves16,Silver17,McKinney20,Senior20,Bellemare20,Ecoffet21,Saggio21,WillettText21,Slonim21,Mirhoseini21,Lu21,Warnat21} but many publications do not report the post-selection stage at al, with few exceptions \cite{Graves16}. 

\end{description}

Below, this author will argue that  (A) suffers from human Post-Selections and (B) suffers from machine Post-Selections.  While Cresceptron, the first deep learning for a 3D world, generates a single large network, it showed an impressive generalization power due to the use of the nearest neighbor scheme at every layer of an automatically generated deep network.  This author will argue that Post-Selections in (A) and (B) suffer from weak generalizations (due to three types of lucks to be discussed below)
and did not count the cost of training multiple networks many of which were not reported. 

By the way, {\em genetic algorithms} offer another approach to such network learning. These algorithms study changes in genomes across different life generations.  However, many genetic algorithms do not deal with lifetime development \cite{WengScience,McClelland07}. We argue that handcrafting functions of a genome as a Developmental Program (DP) seems to be a clean and tractable problem, which avoids the extremely high, cost of evolution on DP.   Many genetic algorithms further suffer from the PSUTS problems, since
they often use test sets as training sets (i.e., vanished tests) as explained below.  

Marvin Minsky \cite{Minsky91} among many scholars complained that neural networks are ``scruffy'' or ``black boxes''.   This problem is not addressed holistically until the framework of Emergent Turing Machine was introduced \cite{WengIJIS15} into Developmental Networks (DNs) by the Developmental School discussed below.   A lack of Emergent Turing Machine mechanisms or being ``scruffy'' in sample fitting appears to be the main cause of PSUTS in traditional neural networks trained by human feature-handpicking or error-backprop  methods.

\subsection{Developmental School}
\label{SE:DevelopmentalSchool}

The main thrust of the Developmental School, formally presented 2001 by Weng and six co-authors \cite{WengScience} is the task-nonspecificity for lifetime development, known as Developmental Programs (DPs) that simulate the functions of genome without simulating the genome encoding.   Although a DP generates a neural network, a DP is very different from a conventional neural network in the evaluation of performance across each life---all errors from the inception time $0$ of each life is recorded and reported up to each frame time $t>0$, as explained further below.

The first developmental program seems to be the Cresceptron by Weng et al.~\cite{WengIJCNN92,WengICCV93,WengCresIJCV97} which appears to be, as far as the author is aware,  the first deep-learning Convolutional Neural Network (CNN) for a 3D world.  As explained in  \cite{WengPlagiarismFaceBook2020,WengPlagiarismYouTube2020} other well-known CNNs for 3D recognition, although they do not use a generative DP, followed many key ideas of Cresceptron.  Cresceptron seems to be the first incremental neural network whose evaluation of performance is across its entire ``life'' and only one network was generated (developed) from the given training data set.   

Cresceptron did not deal with time.   A developmental approach that deals with both space and time in a unfired fashion using a neural network started from Developmental Networks
(DNs) \cite{WengWhy11} whose experimental embodiments range from Where-What Networks 1 (WWN-1) \cite{Ji08} to WWN-9 \cite{Guo15}.  The DNs went beyond vision problems to attack general AI problems including vision, audition, and natural language acquisition as emergent Turing machines \cite{WengIJIS15}.

DNs overcame the limitations of the framewise mapping in Eq.~\eqref{EQ:XLtoLmapping} by dealing with lifetime mapping:
 \begin{equation}
f: X(t-1)\times Z(t-1) \mapsto  Z(t), t=1, 2, ... 
\label{EQ:XZmapping}
\end{equation}
where $X(t)$ and $Z(t)$ are the sensory input space and motor space, respectively, and $\times$ denotes the Cartesian product of sets.  A fundamental difference between Eq.~\eqref{EQ:XLtoLmapping} and Eq.~\eqref{EQ:XZmapping} is that in the latter the $Z$ space contains 
exclusively emergent vectors, instead of any symbols, so that the actions/states are incrementally taught and learned across a lifetime. 

As we will see in Section~\ref{SE:DevError}, all the errors occurred during any time of each life is 
recorded and taken into account in the performance evaluation.   

It is important to extend Eq.~\eqref{EQ:XZmapping} to include the hidden area $Y$ that generates international (hidden) representations.  
To model how $Y$-to-$Y$ connections
enable something similar to higher and dynamic order in Markov models (but not symbolic), the above lifetime mapping is extended to:
\begin{equation}
f: X(t-1)\times Y(t-1) \times Z(t-1) \mapsto  Y(t) \times Z(t), t=1, 2, ... 
\label{EQ:XYZmapping}
\end{equation} 

Note that $Z(t-1)$ here is extremely important since it corresponds to the state of a Turing machine.  Namely, all the errors occurred during any time of each life is 
recorded and taken into account in the performance evaluation.  

Different from the static symbols in the symbolic school and the space of class labels $L$ of static symbols in Eq.~\eqref{EQ:XLtoLmapping} of the connectionist school, the space $Z(t)$  of numeric vectors of the developmental school is free from symbols.
Therefore, these states/actions are directly teachable or self-generative, inspired by brains \cite{FellemanVanEssen91,Super76,Thoroughman,Rizzolatti87,Moore03,Thoroughman,Iverson10}.  This new symbol-free formulation is necessary to model not only brain's spatial processing \cite{WengSpace12}
and temporal processing \cite{WengTime}, but also Autonomous Programming for General Purposes (APFPG)  \cite{WengIJHR2020}.    Based on the APFGP capability, we open the door 
towards the next step---conscious learning \cite{WengCAI-ICDL20}---learning while being partially and increasingly conscious.
By {\em conscious learning}, we do not mean ``open-skulledly'' handcrafting general-purpose consciousness,
which is probably too complicated to handcraft.   But instead we enable fully autonomous machine learning while machines being partially conscious---autonomously learn more sophisticated consciousness skills using their partial, earlier, and simpler conscious skills across the lifetime.

This is the journal archieval version of the earlier conference papers \cite{WengPSUTS21,WengPSUTS-ICDL21} with significantly refined additional material and analysis.

In the following, we will discuss Post-Selection in Section~\ref{SE:post}.   Section~\ref{SE:WhyNeed} addresses why error-backprop  algorithms suffer from severe local minima problems.   
Section~\ref{SE:WhyNotNeed} explains how a Developmental Network solves the local minima problems, since only one network is needed for each life and the evaluation of performance across the entire life.  Section~\ref{SE:Exp} discusses experiments.  Section~\ref{SE:conclusions} provides concluding remarks.

\section{Post-Selections}
\label{SE:post}

AI has made impressive progress, gained much visibility, and attracted the attention of many government officials.   However, there are protocol flaws that have resulted in misleading results.  

First, let us consider three learning conditions that any fair comparisons of AI methods should take into account. 

\subsection{The Three Learning  Conditions}
\label{SE:Three}
Many AI methods were evaluated without considering how much computational recourses are 
necessary for the development of a reported system.   Thus, comparisons about the performance of the system have been tilted toward competitions about how much resources a group has at its disposal, regardless how many networks have been trained and discarded, and how much time the training takes.  

Here we explicitly define the Three Learning Conditions for development of an AI system:

\begin{definition}[The Three Learning  Conditions]
\label{DF:Conditions}
The Three Learning  Conditions for developing an AI system are: (1) a set of restrictions of learning framework, including whether task-specific or task-nonspecific, batch learning or incremental learning, and the body's sensors and effectors; (2) a training experience and (3) a limited amount of computational resources including the number of hidden neurons. 
\end{definition}

The competing standard of the ImageNet competitions \cite{Russakovsky15} did not include any of these three conditions.  The AIML Contests \cite{WengAAAIFS18} considered all the three in performance evaluation.   In the following Subsection, we discuss why task-nonspecificity and incremental mode should be considered in any comparisons. 

\subsection{Task-specific vs. Task-nonspecific}

A task-specific learning approach learns less because much is handcrafted by a human according to the given task. Furthermore, a task-specific method is brittle.

In Condition (1) of the Three Learning Conditions, the task-nonspecific learning paradigm is significantly different from the task-specific traditional AI paradigm as explained in Weng et al. 2001 \cite{WengScience}.   In a task-specific paradigm, the system developer is given a task e.g., constructing a driverless car.   Then, it is a human programmer who chooses a world model, such as a model of lane edges.  Next, he picks an algorithm based on this world model, e.g., the well-known Hough transform algorithm \cite{Ballard,Shapiro01} for line detection which makes every pixel that is detected as edge cast votes for lines of all possible orientations $o$ and distances $d$ from the origin that go through the pixel.  Then
the top-two ``peaks'' of line parameters $(o, d)$ that have received the highest votes are adopted to declare a line detected from the image.   Here ``edges'' and ``lane lines" are two symbolic concepts picked up by the programmer.  Such systems will fail when lanes are unclear or totally disappear due to weather or road conditions, leading to a brittle system.  Human brains appear to be more resilient.

In contrast, a task-nonspecific approach \cite{WengScience} not only avoids any symbolic model, but also does not requires that a task is given.   The desirable actions at any time are taught, tried, and recalled automatically by the learner based on system's learned context $q$ \cite{WengNAI2ed2019} that includes automatically figured-out goal and state, as well as the current input.  The mapping function $f(\z,\x)=\z'$, representing the symbolic mapping $f_s(q, \sigma)=q'$, 
corresponds to a finite automaton.  Weng has proven that the control of any Turing machine is a finite automaton \cite{WengIJIS15}.  Thus, this framework is of general-purposes in the sense of universal Turing machines.  Any universal Turing machine is of general purposes, because it can read any program written for any purposes and run it for the purposes.  Any neural network that learns a universal Turing machine 
become of general purposes in the sense of any programs, not just in the sense of any mappings like that in Eq.~\eqref{EQ:XtoLmapping}.  Thanks to the absence of any world model, such as lanes, this task-nonspecific approach has a potential to be more robust than a world-model based approach.   The task-nonspecific approach typically uses a neural network to learn because the need to learn vector mapping function $f(\z,\x)=\z'$.  We will discuss internal response vector $\y$ in Section~\ref{SE:WhyNotNeed} but task-nonspecificity holds true without $\y$.

\subsection{Batch vs. Incremental Learning Modes}

Neural network learning for the mapping $f$ has two learning modes, batch learning and incremental learning. 

With batch learning, a human first collects a set $D$ of data (e.g., images) and then labels each datum with a desirable output (e.g., command of navigation or class label).  A neural network is trained to approximate a mapping $f$ in Eq. ~\eqref{EQ:XtoLmapping} or Eq.~\eqref{EQ:XLtoLmapping}.
Many batch-learning projects use an error-backprop  method \cite{Krizhevsky12,Karpathy14,LeCun15} which uses a gradient-based method to find a local minimum in error. 

As we will discuss in Section~\ref{SE:WhyNeed}, the gradient in the error-backprop  method does not contain key information of many other data if the learning mode is incremental.  Thus, error-backprop  on a large data set does poorly using a purely incremental learning mode.  Many used a block-incremental learning mode
which suffers from the big data flaw in Theorem~\ref{TM:BDF} below. 

In contrast, all developmental methods cited here use incremental learning mode for long lifetimes, using a closed-form solution to the global lifetime optimization.   The competition among neurons guarantees that the winner is the most appropriate neuron whose memory corresponds to the current working memory \cite{WengLCA09}. 

However, the batch and incremental learning modes are not capability-equivalent \cite{WengLCA09}.   The former requires all
sensory inputs are available at a batch, independent of the corresponding actions.    Therefore, the former
is easier and also incorrect according to sensorimotor recurrence.  By  sensorimotor recurrence, we mean that sensory inputs and 
motor outputs are mutually dependent on each other in such a recurrent way that off-line collection of
inputs are technically flawed.   We have the following theorem:
\begin{theorem}[Big Data Flaw]
\label{TM:BDF}
All static ``big data'' sets used by machine learning violate the {\em sensorimotor recurrence} property of the real world.
\end{theorem}
\begin{IEEEproof}
A learning agent at time $t-1$, as shown in Eqs.~\eqref{EQ:XZmapping} and ~\eqref{EQ:XYZmapping} does not have the next sensory input from $X(t)$ available before the corresponding actions in $Z(t-1)$ are generated and output, since the sensory input in $X(t)$ varies according to the agent actions  in $Z(t-1)$.  As an example, turning head left or right will result in a different image sensed.   Therefore, all static ``big data'' sets violate the sensorimotor recurrence. 
\end{IEEEproof}

One may say that classifications of static images are fine.  We do not agree, because even when a human
(or machine) is looking at a static image, he uses attention (e.g., context-based saccades)which is a sequence of actions.
Each saccade results in a different fovea image. 

Therefore, incremental learning is necessary for the {\em sensorimotor recurrence}.
All batch-training methods use a static set of training data and, therefore, are inappropriate for any of them to claim near-human performance since the two learning problems are different.  This leads to the following theorem.

Consider a hierarchy of levels of object types, such as nails, fingers, palms, hands, arms, limbs, torsos, human bodies, etc.  Because vision requires a high level $l$ to understand natural scenes with abstraction of parts with invariances (e.g., all fingers of different scales, looks, and at different locations), each child needs an open-ended world to learn to learn rules (e.g., finger-parts and hand-whole) instead of simple-minded pattern recognition of sensory images.  

\begin{theorem}[Nonscalability of Big Data without abstraction]
\label{TM:Nonscalability}
\label{TM:abs}
All static ``big data'' sets used by machine learning are nonscalable if they are treated as pattern recognition without rule abstractions.
\end{theorem}
\begin{IEEEproof}
Suppose that a static data set $D$ has shown the presence of $k$ feature types defined at level 1 (e.g., edge pixels are a type).  Suppose a combination of $k>1$ feature types to level $l+1$ type (e.g., straight line type is from multiple edge pixels) is defined from $k$ types of feature types at level $l$, $l=1, 2, ... $.  The number of samples for a $l$-level feature type requires at least $k^l$ observations to discover all necessary within-type equivalence (e.g., logic OR is at $l=2$ with $k=2$ logic features at $l=1$, thus without  rule abstraction (e.g., parts and whole), it requires $k^l=2^2=4$ observations, corresponding to 4 rows in the truth table of logic OR).  Since $f(l)=k^l$ is an exponential function in $l$, $k^l$ quickly exceeds any fixed number of observations in the static data set $D$.  
\end{IEEEproof}

Rule abstractions deal with invariances.  For example, a ``what'' concept is ``where''-invariant and a ``where'' concept is ``what''-invariant, as explained in \cite{WengSpace12,WengIEEE-IS2014}.  

Section~\ref{SE:WhyNotNeed} discusses an optimal framework through which such abstractions can take place
from learning simple rules during early life that enable learning of more complex rules during later life---called scaffolding \cite{Wood76}.  

 Theorem~\ref{TM:Nonscalability} leads to two observations on {\bf data fitting on a static data set}:

{\bf Observation 1:}  Any {\bf data fitting on a static data set} without learning invariant concepts are nonscalable, including the $n$-fold cross-validation discussed below.  Unfortunately, {\bf data fitting on a static data set} is a norm in all ImageNet Contests \cite{Karpathy14}. Namely, the remaining subsections in this section analyze approaches that are nonscalable.  For example, computer vision is not a ``one-shot'' pattern classification problem as argued by Li Fei-Fei et al. \cite{Fei-Fei06} (which was questioned in PubMed without responses), but rather a spatiotemporal problem to learn various invariant concepts present in cluttered natural scenes through autonomous attention saccades, as explained further in Observation 2.

{\bf Observation 2:} Learning invariant concepts seem nonscalable for any {\bf data fitting on a static data set} either, because there are too many images to be labeled by hand (e.g., all pixel locations) \cite{WengSpace12,WengIEEE-IS2014}.   Like a human baby, any scalable machine learning methods must be conscious through which the machine learner must consciously guess concepts (i.e., not just active learning \cite{BurrActiveLearning98}) (e.g., an object type) and verify their invariance rules (e.g., the where-invariance of a what concept).   The state-based transfer in Theorem 8 of \cite{WengTime} explains how each concept state reduces the number of samples to be learned from an exponential $k^l$ down to only $kl$ (see Fig. 6 of \cite{WengTime} for intuition where $k=10$ and $l=3$).   Thus, Section~\ref{SE:WhyNotNeed} not only addresses the non-scalability problems in this section, but is also necessary for conscious learning whose theory was recently published in Weng 2020 
\cite{WengCAI-ICDL20} with some single-sensory-modality experimental results, but animal-level conscious robots that are
multi-sensory and multi-motor have not yet been demonstrated.  The availability of real-time learning brain-chip is a current bottleneck.   

\subsection{Fitting, validation and test errors}

Given an available data set $D$, $D$ is divided by a partition $P=(T, V, T')$ into three mutually disjoint sets, a training set $T$, a validation set $V$, and a test set $T'$ so that
\begin{equation}
D=T\cup V \cup T'.
\label{EQ:disjoint}
\end{equation}
Two sets are disjoint if they do not share any elements.   
The validation set is possessed by the trainer, the test set should not be possessed by the trainer since the test should be conducted by an independent agency.  Otherwise, $V$ and $T'$ become equivalent. 

As we will see in Section~\ref{SE:WhyNeed}, given any architecture 
parameter vector $\a_i$, it is unlikely that a single network initialized by a set of random weight vectors can 
result in an acceptable error rate on the training set, called fitting error, that the error-backprop  training intends to minimize locally.
That is how the multiple sets of random weight vectors come in.  For $k$ architecture vectors $\a_i$, $i=1, 2, ... k$ and $n$ sets of random initial weight vectors $\w_j$, the error back-prop training results in $kn$ networks 
\[
\{N(\a_i, \w_j) \;|\; i=1, 2, ... , k, j=1, 2, ..., n\} .
\]
Error-backprop locally and numerically minimizes the fitting error $f_{i,j}$ on the training set $T$.   

\cite{Graves16} seems to have mentioned $n=20$.  \cite{Krizhevsky17} did not give $n$ but seems to have mentioned 60 million parameters which probably means each $\w_i$ and each $\a_j$ combined to be of
60 million dimensional.   Using the above example of $k=3^{10}=59049$, $kn \approx1$M networks must be trained, a huge number that requires a lot of computational resources to do number crunching and a lot of manpower to manually tune the range of hyper-parameters. 

\begin{definition}[Distribution of fitting, validation and test errors]
The distributions of all $kn$ trained networks' fitting errors $\{f_{ij}\}$, validation errors $\{e_{ij}\}$, and test errors $\{e'_{ij}\}$, $i=1, 2, ... k$, $j=1, 2, ... n$ are random distributions depending on a specific data set $D$ and
its partition $P=(T, V, T')$.  The difference between a validation error and a test error is that the former is computed from the same group using 
a group-possessed validation set $V$ but the latter is computed by an independent agency using a 
group-unknown test set $T'$.
\end{definition}

We define a simple system that is easy to understand for our discussion to follow. 

\begin{definition}[Nearest neighbor classifiers with a confidence threshold]
\label{DF:NN}
Define a network stores the entire training set $T$.  Suppose the input $x$ matches the nearest sample $s$ in $T$.  If the distance between $x$ and $s$ is not larger than a confidence threshold $d$ (a hyper-parameter), then 
the network outputs the associated label of the nearest sample $s$. 
Otherwise, the system outputs ``unknown''.
\end{definition}  

Namely, this system uses a lot of resources for over-fitting.  It gives up if the distance is larger than $d$, but has a perfect fitting error (zero) for any positive $d$.

A neural network architecture has a set of hyper parameters represented by a vector $\a$, where each component corresponds a scalar parameter, such as convolution kernel sizes and stride values at each level of a deep hierarchy, the neuronal learning rate, and the neuronal learning momentum value, etc.  Let $k$ be a finite number of grid points along which such architecture parameter vectors need to be tried,  $A = \{ \a_i \;|\; i=1, 2, ..., k\}$.  Suppose there are 10 scalar parameters in each vector $\a_i$.  For each scalar parameter $x$ of the 10 hyper parameters, 
we need to validate the sensitivity of the system error to $x$.   With uncertainty of $x$, we estimate its initial value as the mean $\bar{x}$, positively perturbed estimate $\bar{x}+\sigma_x$ ($\sigma$ is the estimated standard deviation of $x$), and negatively perturbed estimate $\bar{x}-\sigma_x$.  If each scalar hyper parameter has three values to tray in this way, there are a total of $k=3^{10}=59049$ architecture parameter vectors to try, a very large number.   For example, the initial threshold $\bar{d}$ in the nearest neighbor classifier can be estimated by the average of nearest distance between a sample in $V$ and the nearest neighbor in $T$ and the $\sigma_d$ be estimated by the standard deviation of these nearest distances. 

Let us define the Post-Selection.  Suppose that the trainer is first aware of the validation sets (or the test sets).   

\begin{definition}[Post selection]
A human programmer trains multiple systems 
using the training set $T$.   After these systems have been trained, he post-selects a system by searching, manually or assisted by computers, among trained systems based on the validation set $V$ (or the test set $T'$).  This is called Post-Selection---selection of one network from multiple trained and verified (or tested) networks.
\end{definition} 

Obviously, a post-selection wastes all trained systems except the selected one.   As we will see next, a system from the post-selection tends to have a weak generalization power. 

First, consider Post-Selection Using Validation Sets (PSUVS):

\subsection{PSUVS}

A Machine PSUVS is defined as follows:
 If the test set $T'$ is not available, suppose the validation error of $N(\a_i, \w_j)$ is $e_{i, j}$ on the validation set $V$, find the luckiest network $N(\a_{i^*}, \w_{j^*})$ so that it reaches {\em the error of the luckiest-architecture and the luckiest initial weights from Post-Selection on Validation Sets}: 
\begin{equation}
e_{i^*,j^*} = \min_{1\le i \le k} \min_{1\le j \le n} e_{i,j}
\label{EQ:e}
\end{equation}
and report only the performance $e_{i^*,j^*}$ but not the performances of other remaining $kn-1$ trained neural networks.  

Similarly, a human PSUVS is a procedure wherein a human selects a system from multiple trained systems 
for $\{ e_{i,j} \}$ using human visual inspection of internal representations of the system and their validation errors.  

\subsection{Cross-Validation}

The above PSUVS is an absence of cross-validation \cite{JainDubes}.   Originally, the cross-validation is meant to mitigate an unfair luck 
in a partition of the dataset $D$ into a training set $T$ and a test set $T'$ (empty validation set).  For example, an unfair luck 
is such that every point in the test set $T'$ is well surrounded by points in the training set $T$.  But such a luck is hardly true in reality. 

To reduce the bias of such a luck, an $n$-fold cross-validation protocol, $n\ge 2$, divides the data 
set $D$ into $n$ subsets of same size and conducts $n$ experiments.  The term ``cross'' refers to 
switching the roles of training and testing data.  In the $i$-th experiment, 
the $i$-th subset is left out as the test set and the remaining $n-1$ folds of data form the training set.
Thus, the cross-validation protocol conducts $n$ experiments, for $i=1, ... , n$, to obtain $n$ errors, $e_1, e_2, ... , e_n$.   The {\em cross-validated error} is defined as the average of errors from the $n$ tests, to filter out the partition lucks:
\begin{equation}
\bar{e} = \frac{1}{n} \sum_{i=1}^{n} e_i 
\label{EQ:cross}
\end{equation}
as well as the distribution of errors $\{ e_j\}$. 

The $n$ different numbers here shows a distribution show a distribution $\{ e_j\}$ to 
indicate how sensitive the error is to lucks, such as the number of partition pairs between a training set and a validation/test set, the number of tried random seeds for initializing network weights, the number of tried hyper-parameter vectors, or a combination thereof.   The larger the $n$, the better the estimated 
standard deviation of $\{ e_j\}$.  

\subsection{Types of lucks in a Neural Network}

 In a neural network, there are at least three kinds of lucks:

{\bf Type-1 order lucks}: The luck in a partition $P_i$ into a training set $T_i$ and a test set $T'_i$ from a data set $D$ resulting in test error $e_i$, $i=1,2, ... n$.   Different partitions correspond to different luck outcomes.  This kind of outcome variation results in a variation of performance from different outcomes.   Conventionally, this type of lucks is filtered out by cross-validation (e.g., $n$-fold cross-validation) as well as reporting the deviation of $\{e_i\}$ during the cross-validation.  However, such cross-validation and deviation have hardly published for neural networks and reported.   The smaller the average $\bar{e}$ of $\{e_i\}$, the more accurate the trained network is; the smaller the standard deviation of $\{e_i\}$, the more trustable the average error $\bar{e}$ is. 

{\bf Type-2 weights lucks}: As we will discuss below, weights specify the role assignment for 
all the neurons in the neural network.  A random seed value determines the initialization of  a pseudo-random number generator, which gives initial weights $\w_i$ for a neural network $N(\w_i)$, resulting in a test error $e_i$, $i=1,2, ... n$, after training of these $n$ networks and testing on $T'$.   It is unknown that such a 
luck will be carried over to a new test set $T''$ that is outside the data set $D$ but was drawn from the same distribution of $S$.  Because a neural network might not capture the internal rules of the training set $T$, this paper argues that a statistical validation of the reported error should be performed by reporting the distribution of $\{e_i | i=1, 2, ... n\}$, where $e_i$ is from a different initial weight vector $\w_i$.   For example, Krizhevsky et al. \cite{Krizhevsky17} reported 60 million parameters, mostly in $\w_i$ but only the luckiest  $e_i$ was reported.  The smaller the average $\bar{e}$ of $\{e_i\}$, the more accurate the trained network is; the smaller the standard deviation 
$\sigma$ of $\{e_i\}$, the less sensitive the trained neural network is to the initial weights and thus the accuracy is more trustable for real applications.  For i.i.d. (identically independently distributed) errors, 
we can expect that doubling the number $n$ will reduce the expected variance of $\bar{e}$ by a factor $1/\sqrt{2}$., since the expected variance of $n$ random numbers is about $\sigma^2/n$.
 
{\bf Type-3 architecture lucks}: The initial hyper-parameter vector $\a_j$ of the neural network gives an error $e_j$, $j=1,2, ... k$.   Because such a luck of $\a_j$ might not capture the internal rules of the training set $T_j$, this paper argues that a statistical validation of the reported error estimate should be performed and the distribution of $\{e_j\}$ be reported.    In our above example, the number of distinct hyper-parameter vectors to be tried is $k=3^{10}=59049$.   The smaller the average $\bar{e}$ of $\{e_j\}$, the more accurate the trained network is; the smaller the sample variance of $\{e_j\}$, the more trustable $\bar{e}$ is, namely, 
the average error $\bar{e}$ is less sensitive to the initial hyper-parameters of the network.    For example, 
the threshold $d$ of the nearest neighbor classifier in Definition~\ref{DF:NN} might result in a large deviation. A good way is to reduce the manual selection nature of such hyper-parameters.  For example, all hyper-parameters are adaptively adjusted from the initial hyper-parameters that are further automatically computed 
from system resources, e.g., the resolution of a camera, the total number of available neurons, and the firing age of each neuron \cite{WengLCA09}. 

For notation clarity in the discussion that follows, index $j$ is used in Type 3 to distinguish index $i$ in type 2, but the above three types of lucks are all different. 

Let us discuss the case of a developmental network, such as Cresceptron \cite{WengCresIJCV97} and DN \cite{WengIJIS15}.  Type-1 cross-validation is not needed because of reporting of a lifetime error.
In other words, errors of all new tests in each life are taken into account throughout the lifetime.
Type-2 validation is not needed because all different random weights $\w_i$ leads to the function-equivalent  neural network under certain conditions.  For example, in top-$k$ competition, with $k=1$ different 
$\w_i$ give the exactly the same neural network and with $k>1$ different 
$\w_i$ give almost the same neural network.   The distribution of lifetime errors $\{e_i\}$ is expected to have a negligible deviation across different initial weight vectors $\w_i$, given the same Three Learning Conditions.
Type-3 validation might be useful but is expected to be negligible since the most obvious parameters such as learning rate and momentum of learning rate is automatically and optimally determined by each neuron, 
not handcrafted, as in LCA \cite{WengLCA09}.   The synaptic maintenance automatically adjusts all receptive fields \cite{Wang11,GuoIJCNN14} so that the neural network performance is not sensitive to the initial hyper-parameters. 
 
In contract, a batch-trained neural network typically uses a Post-Selection to pick the luckiest network without cross-validation for either of the above three types of lucks, e.g., in ImageNet Contest \cite{Russakovsky15}.  Namely, errors occurred during batch training of the network before the network is finalized and how long the training takes are not reported.  Below, Fig.~\ref{FG:CNN-vs-DN} will show a huge difference between the luckiest CNN with error-backprop  and the optimal DN.  Many researchers have claimed error-backprop works without providing much-needed three types of validations.   

Next, let us discuss Types 2 and 3 validations which are new for neural networks but hardly done.

\subsection{Post-Selection with Types 2 and 3 Average-Validations}
Type-1 cross-validation should be nested inside the Types 2 and 3 validations, but this triple-nested protocol could be too computationally expensive.   Below, we delay Type-1 cross-validation till after Type 2 and Type 3 validations.  

Assume that we use $n$ random weight vectors $\w_i$ and $k$ grid-search hyper parameters $\a_j$.
Each combination of $\w_i$ and $\a_j$ gives an error $e_{i,j}$ from the corresponding validation set.
To reduce the effect of such a luck for each 
vector $\w_i$, an average of $e_{i,j}$ over $n$ values of $i$ should be used instead of the minimum in Eq.~\eqref{EQ:e}.  This leads to 
{\em the 
random-weights validated error for the luckiest architecture} from PSUVS: 
\begin{equation}
\a^* = \arg\min_{1\le j \le k} \frac{1}{n} \sum_{i=1}^{n} e_{i,j} .
\label{EQ:ecrossn}
\end{equation}
We dropped the term ``cross'' because this validation examines other random seeds without switching the roles between training and testing.   

Similarly, we define {\em the hyper-parameter validated luckiest initial weights} from PSUVS:
\begin{equation}
\w^* = \arg \min_{1\le i  \le n} \frac{1}{k} \sum_{j=1}^{k} e_{i,j} .
\label{EQ:ecrossk}
\end{equation}
We dropped the term ``cross'' for the same reason.

From a statistical point of view, the initial hyper parameter vector $\a^*$ and the random initial weights $\w^*$ validated above through averages should be more robust in real applications than those without average-validation in Eq.~\eqref{EQ:e}.

For both the luckiest $\a^*$ and $\w^*$, the standard deviation under $\min$ should be reported to show how sensitive the reported performance is to the validation process. If the variation is large, the corresponding network is not very trustable in practice. 

We also need to be aware of another protocol flaw:  Random seeds and hyper parameters are all coupled. 
Under such a coupling, Type 2 validation seems unnecessary with $n=1$ but the search of the luckiest weights is embedded into the search for the luckiest hyper-parameter vector where each hyper parameter vector uses a different seed.  Similarly, Type-3 validation seems unnecessary with $k=1$ but the search of the luckiest hyper-parameter vector is embedded into the search for the luckiest weights, where each 
random seed uses a different hyper parameter vector.   

Since a PSUVS procedure picks the best system based on the errors on the validation set, the resulting system might not do well on the test sets because doing well on a validation set does not guarantee doing well on a test set.  Typically, due to a very large number of samples, availability of validation sets and unavailability of test sets in a properly managed contest, principles of Post-Selection should cause 
the validation error rate to be smaller than the test error rate.   (However, in Table 2 of \cite{Krizhevsky17}, the test error rate is smaller than the validation error for 7CNNs, causing a reasonable suspicion that PSUTS could be used instead of PSUVS.)  

The following subsection discusses the luckiest network with the luckiest hyper-parameter vector $\a^*$ and the luckiest initial weights $\w^*$.

\subsection{The Luckiest Network from a Validation Set}

Many people may ask:  Are there any technical flaws in at least PSUVS, since it does not use the test sets?  We analyze the luckiest network in this section and reach a conclusion that any post-selection is technically flawed and results in misleading results, including both PSUVS and PSUTS.  However, in general, Type-1 cross-validation is to filter out lucks in data partition that a typical user does not have during a deployment of the
method.   Namely, it is a severe technical and protocol flaw in reporting only the luckiest network, regardless the post-selection uses validation sets or test sets.   

This conclusion has a great impact on evolutional methods that often report only the luckiest network, instead of those of all networks in a population.
Namely, the performances of all individual networks in an evolutionary generation should be reported. 

For simplicity, we assume that the space $S$, from which random samples in $D$ are drawn, is static.  Our conclusions here can be readily extended to a time varying $D$ but the technical flaws are even worse. 

From the sample space $S$, randomly draw a data set $D$.  $D$ is partitioned into three mutually disjoint sets, 
training set $T$, validation set $V$ and test set $T'$, so that Eq.~\eqref{EQ:disjoint} holds true.   For 
realistic applications, we should assume that $T$, $V$ and $T'$ are mutually independently drawn from $S$ so that $T$, $V$ and $T'$ are mutually independent.  Identically independently distributed (i.i.d.) is 
a sufficient condition, but we do not need such a restrictive condition because temporal-dependency 
often occurs in lifetime development.   Namely, we only need that any three vectors from $T$, $V$ and $T'$, respectively, are mutually independent. 

Using the training set $T$, one trains $kn$ networks, where $k$ and $n$ are the number of hyper-parameter vectors $\a$'s and random weight vectors $\w$'s, using a training algorithm (e.g., error-backprop ),
\begin{equation}
N(\a_i, \w_j) \leftarrow f_{\a_i, \w_j} ( T).
\label{EQ:trainingalg}
\end{equation}
This is like a teacher trains 
$kn$ students in a class.  The teacher knows that the fitting error on $T$ does not predict the validation error well, due to the possibility of over fitting.  One extreme example is the above nearest-neighbor classifier 
with confidence $d=0$.   

The teacher then tests each $N(\a_i, \w_j) $
on the validation set $V$ to get $e_{i,j}$.   This is like the teacher observes the performance of $kn$ networks in a mock exam.   

The teacher then post-selects and reports only the luckiest network $N(\a_{i^*}, \w_{j^*})$ whose validation error $e_{i^*, j^*} $ is minimum in Eq.~\eqref{EQ:e}.   This is like the teacher colludes with the Educational Test Service (ETS) so that the ETS only reports the luckiest network but not all remaining $kn-1$ networks to cover up. 

\subsection{Luckiest Network with Type-1 Cross-Validation}
Suppose that a user has bought the luckiest network $N(\a_{i^*}, \w_{j^*})$ and test on his new test data $T'$ randomly drawn from $S$, independent of $T$ and $V$.   The luckiest 
network $N(\a_{i^*}, \w_{j^*})$ that reached the minimum error rate in $V$ does not mean that it reaches
the minimum error rate in $T'$.  Because $T'$ is independent of $T$ and $V$, and $\a_{i^*}, \w_{j^*}$ are luckiest on a particular pair $(T, V)$ only, we need to compute the expected error rate of $N(\a_{i^*}, \w_{j^*})$ on $T'$.

\begin{theorem}[Type-1 cross-validation of the luckiest]
The luckiest network on validation set gives an error rate that is approximately the average error in Type-1 cross-validation, supposing that, in an $n$-fold cross validation, $n$ folds of data are drawn i.i.d. (independently and identically distributed) among folds from data set $D$, but individual samples inside each fold do not need to be i.i.d.
\end{theorem}
\begin{IEEEproof}
Let $F$ denote the event that both the training set and validation set are from a fixed data set $D$ from $S$. Consider in a real application, $n$ tests were conducted on the luckiest network $N(\a_{i^*}, \w_{j^*})$ using $T'_i$, $i=1, 2, ... , n$, where each partition $P_i=(T_i, V_i, T'_i)$ in each of the $i$-th training and test pair is drawn from the real application space $S$.   We 
compute the average error rate from the luckiest network $N(\a_{i^*}, \w_{j^*})$:
\begin{eqnarray}
e(\a_{i^*}, \w_{j^*}) & =&  \frac{1}{n} \sum_{i=1}^{n} e_i (\a_{i^*}, \w_{j^*}; T_i, V_i , T'_i)  \nonumber \\
 & \approx & \frac{1}{n} \sum_{i=1}^{n} e_i (\a_{i^*}, \w_{j^*}; T_i, V_i , T'_i |F ) 
 \label{EQ:addF}
\end{eqnarray}
where the term $e_i (\a_{i^*}, \w_{j^*}; T_i, V_i, T'_i)$ means the error of the luckiest network 
using training set $T_i$, validation set $V_i$, and test set $T'_i$, and $e_i (\a_{i^*}, \w_{j^*}; T_i, V_i , T'_i | F )$ means the same but $T_i, V_i, T'_i$ are all from the same $D$ as 
one does in $n$-fold cross-validation.   
\end{IEEEproof}

Note, the $n$-fold i.i.d. is weaker than i.i.d. for all samples.  In practice, i.i.d. is rarely true even for pattern recognition problems, such as image classification due to sequential attention discuss above.    Also note that the left side 
of $\approx$ sign in Eq.~\eqref{EQ:addF} is expected larger because the data $T_i, V_i, T'_i$ on the right side are all from a fixed $D$ but the left side does not have such a restriction.

The above theorem tells us that
the error rate of the luckiest network from a single validation set in PSUVS is misleading without any partition validation.   This is because the error rate is a random function, depending on not only many random initial weights, many hyper parameters, and local lucks of error-backprop, but also a particular partition $(T,V,T')$.   This seems especially true if the data $D$ were made public and overworked during
2010-2014 \cite[p. 213]{Russakovsky15}.

In practice, when we report an error rate $e(\a_{i^*}, \w_{j^*})$ which is always a random number $x$,
depending on how much hand tuning is done, how much computational resources are used for a large-scale search for the random seeds and hyper-parameters, as well as the validation or a lack thereof.   We should also report the distribution of this random number $x$, such as the maximum, 75\%, 50\%, 25\%, and the minimum value of $x$, over multiple training-and-test pairs in cross-validation, random seeds and hyper-parameters.   Otherwise, the 
error rate, if only as a single number $x$, is misleading, since users of this learning method or buyers of the luckiest network do not have the same partition luck. 

Up to now, this author has not found any published papers that report not only 
the luckiest network from error-backprop  but also Type-1, Type-2 and Type-3 validations.   
Many papers do not report the post-selection stage at all \cite{LeCun15,Mnih15,Silver16,Silver17,McKinney20,Senior20,Bellemare20,Ecoffet21,Saggio21,WillettText21,Slonim21,Mirhoseini21,Lu21,Warnat21}, except \cite{Graves16}, let alone whether the reported error is from the validation error $V$ or the test set $T'$.   

Next, we discuss Post-Selections Using Test Set (PSUTS).  There are two kinds of PSUTS, machine PSUTS and human PSUTS.

\subsection{Machine PSUTS}

If the test set $T'$ is available which seems to be true for almost all neural network publications other than competitions, we define Post-Selection Using Test Sets (PSUTS):

A Machine PSUTS is defined as follows:
If the test set $T'$ is available, suppose the test error of $N(\a_i, \w_j)$ is $e'_{i, j}$ on the test set $T'$, find the luckiest network $N(\a_{i^*}, \w_{j^*})$ so that it reaches the minimum, called {\em the error of the luckiest architecture and the luckiest initial weights from Post-Selection on Test Set}: 
\begin{equation}
e'_{i^*,j^*} = \min_{1\le i \le k} \min_{1\le j \le n} e'_{i,j} .
\label{EQ:e'}
\end{equation}
Report only the performance $e'_{i^*,j^*}$ but not the performances of other remaining $kn-1$ trained neural networks.

Imagine that we want to remove lucks in the above expression, by using averages like we did in Eq.~\eqref{EQ:cross} to give {\em the error of the luckiest architecture with validated weights from PSUTS}:
\begin{equation}
\a_{*,j^*} = \arg \min_{1\le j \le k} \frac{1}{n} \sum_{i=1}^{n} e'_{i,j}.
\label{EQ:e'cross}
\end{equation}
But the above error is still flawed since each term under minimization has peeked into test sets.
Instead, it is better to use Eq.~\eqref{EQ:ecrossn} which does not use the test sets.  Of course, the 
test error rate of that in Eq.~\eqref{EQ:ecrossn} tends to be larger than that from Eq.~\eqref{EQ:e'cross}.

A similar discussion can be made for {\em the error of the luckiest initial weights with validated architecture from PSUTS}.  Do not peek into test sets.

There are some variations of Machine PSUTS:  The validation set $V$ or $T'$ are not disjoint with $T$.  If $T=V$, we
call it validation-vanished PSUTS.  If $T=T'$, we called it test-vanished PSUTS.   

In general, the more free parameters a network has, the more likely the network can report an artificially small error as in Eq.~\eqref{EQ:e'}.   That is why we need the computational resource in the Three Learning Conditions. 

Although PSUVS has flaws of post-selection and a lack of three types of validation, the key difference between PSUVS and PSUTS does not guarantee that PSUVS reports a low error rate as PSUTS. 
In fact, it is 
expected that the luckiest network from PSUVS does better on a validation set $V$ than on a test set $T'$ because the Post-Selection did not ``see'' the test set $T'$ but ``saw'' the validation set $V$.
Likewise, it is expected that the luckiest network from PSUTS
does better on the test set $T'$ than on a validation set $V$ because the Post-Selection did not ``see'' the validation set $V$ but ``saw'' the test set $T'$.  In the following paragraph, we discuss that this expectation is reversed in Table 2 of \cite[page 88]{Krizhevsky17}.

In ImageNet Contest 2012, the test sets were released to competition teams over 2.3 months ahead of the output-result submission date.  Although the class labels were not attached to the test sets other than being available indirectly through an  online test server provided by the contest organizers, it was not difficult to ``crack'' a test set by manually hand-labeling the test set.  The first author of \cite{Krizhevsky17} seems not sensitive to the fundamental difference between a validation set and a test set by writing: ``in the remainder of
this paragraph, we use validation and test error rates interchangeably''.  By ``we cannot report test error rates for all the models that we tried'' \cite[page 88]{Krizhevsky17}, there is no evidence to rule out what he meant was the possibly ``cracked'' test set is not necessarily exactly the same as the original test set.  But in Table 2 of \cite[page 88]{Krizhevsky17}, the 7NNs did worse on the validation set (possessed) than the test set (if not ``cracked'' and searched for minimization like in Eq.~\eqref{EQ:e'}).  This reversed our  expectation in the previous paragraph.  Is it an evidence of using PSUTS instead of PSUVS?

Another interesting phenomenon that is consistent with the likely use of PSUTS instead of PSUVS is that the SuperVision Team of ImageNet Contest 2021 did not submit any output results for ``the fine-grained classification task, where algorithms would classify dog photographs into one of 120 dog breeds'' \cite[footnote, p214]{Russakovsky15}.  It appears that cracking ``120 dog breeds'' is harder than cracking ``a list of object categories present in the image'' where the class labels are all available in the provided training sets.  \cite{Krizhevsky17} lacks due transparency about the post-selection stage except that 
Geoffrey Hinton admitted the ``luckiest'' network in his brief PubPeer response to questions raised on PubPeer towards \cite{LeCun15}.

\begin{figure}[tb]
	\centering
	\includegraphics[width=0.6\linewidth]{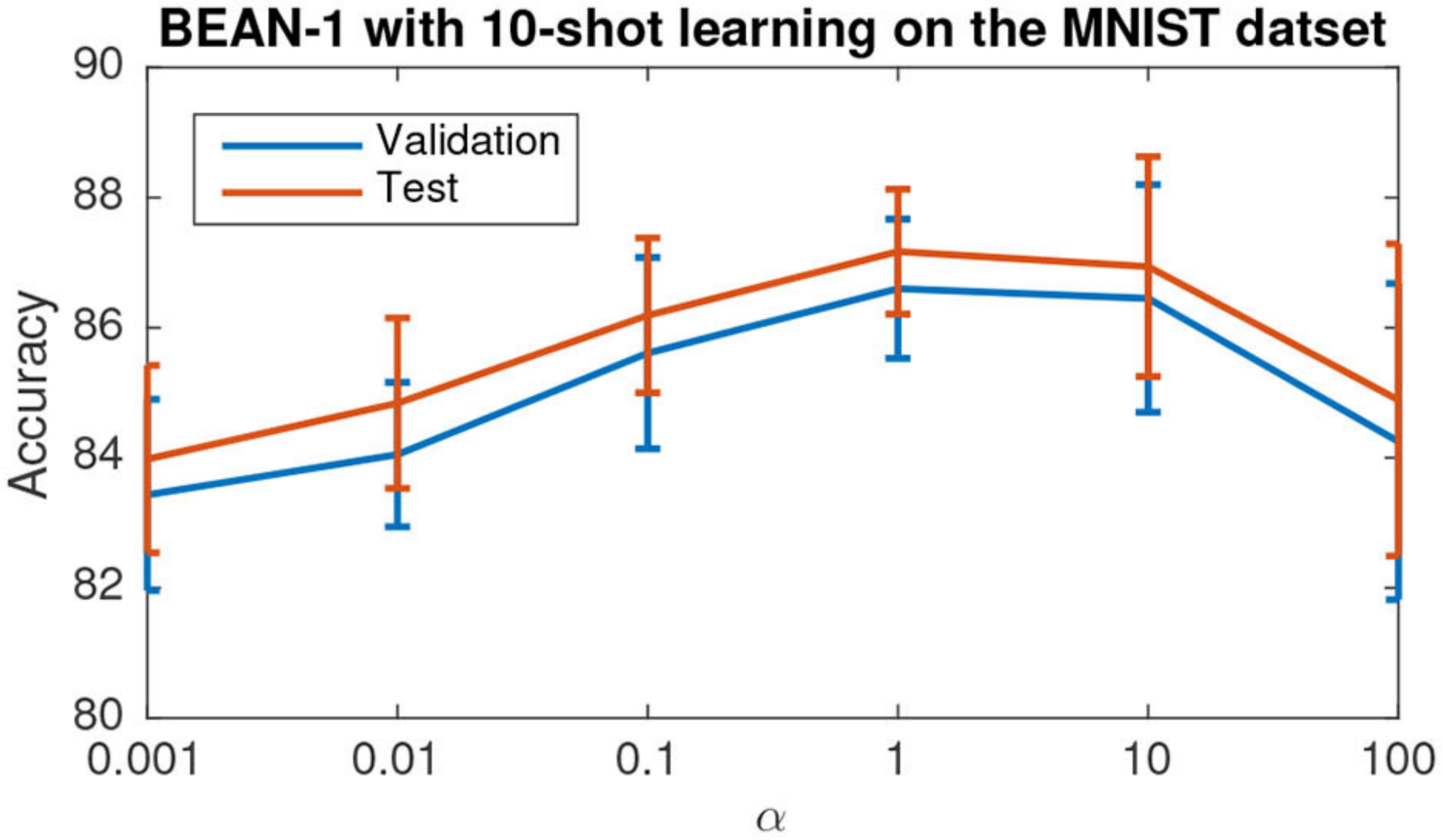}
	\caption{
	The average $\bar{e}$ and the standard deviation of $\{e_i\}$ for different values of a regularization hyper parameter $\alpha$.  Adapted from \cite{GaoBEAN21}}
	\label{FG:BEAN1-PSUTS}
\end{figure}

For more examples, see Fig.~\ref{FG:BEAN1-PSUTS} from \cite[Fig. 7]{GaoBEAN21}, error-backprop consistently results in lower validation accuracies than the test accuracies (about 0.5\% lower compared to 
about 0.1\% lower in \cite{Krizhevsky17}).  Are they other evidences of using PSUTS instead of PSUVS, similar to  \cite{Krizhevsky17}?  The availability of test sets to the programmers in a project seems to be indeed addictive towards PSUTS, away from PSUVS.  The standard deviation around $1\%$ is clashes 
with our Theorem~\ref{TM:lack}.   
Our experience with our own experiments with error-backprop training for CNN indicated that the maximum and the minimum values of the distribution of fitting accuracies are drastically different for different random seeds, with fitting accuracies spreading uniformly between 20\% and 90\%.   Section~\ref{SE:WhyNeed} will discuss why.   If Theorem~\ref{TM:lack} in Section~\ref{SE:WhyNeed} is correct, the deviation bars 
seem too small and the 20 runs in Fig. 7 of \cite{GaoBEAN21} could be the best 20 among many more random-seeds the programmer has tried.  We hope that authors provide the source program. 

\subsection{Implications of PSUTS}

Although the set $\{e_{i,j} | i=1, 2, ... n; j=1, 2, ... k\}$ is large, it is necessary to present some key statistical characteristics of its distribution.  For example, rank all errors in decreasing order, for each type of errors, fitting, validation and test.  Then give the maximum, 75\% (in ranked population), 50\% (median), 25\%, the minimum value, and the standard deviation of these $kn$ values for the fitting errors, validation errors. and test errors, respectively, not just the standard deviation in Fig.~\ref{FG:BEAN1-PSUTS} . Such more complete information of the distribution is critical for the research community to 
see whether error-backprop can indeed avoid local minima in deep learning as some authors claimed.  
Furthermore, such information is also important for the authors to show that the luckiest hyper-parameter 
vector is not just an over fitting to the validation/test set.  Unfortunately, none of \cite{Krizhevsky12,Karpathy14,LeCun15,Mnih15,Graves16,Moravcik17,McKinney20,Bellemare20} reported such distribution characteristics other than the minimum value $e'_{i^*,j^*}$.  

Furthermore, such a use of test sets to post-select networks resembles hiring a larger number $kn$ of random test takers and report only the luckiest $N(\a_{i^*}, \w_{j^*})$ after the grading.  This practice could hardly be acceptable to any test agencies and 
any agencies that will use the test scores for admission purpose since this submitted error $e'_{i^*,j^*}$ misleads due to its lack of validation.  

The error-backprop  training tends to locally fit each network on the training set $T$; while the Post-Selection 
picks the luckiest network with parameter vector $\a_{i^*}$ and initial weights $\w_{j^*}$ that has the best luck on $T'$.  If an unobserved data set $T''$, disjoint with $T'$, $T'\cap T'' = \emptyset$, is observed from the same distribution $S$, the error rate $e''_{i^*,j^*}$ of $N(\a_{i^*}, \w_{j^*})$ is predicted to be significantly higher than $e'_{i^*,j^*}$,
\begin{equation}
e''_{i^*,j^*}  \gg  e'_{i^*,j^*}
\label{EQ:gg}
\end{equation}
because Eq.~\eqref{EQ:e'} depends on the test set $T'$ in the post selection from many networks.  Of course, handling a new test is also challenging for a human student but a human learning involves 
learning invariant rules.   Any PSUTS is a technically flawed protocol.   

PSUTS is tempting especially when test sets are available to the authors of paper.   During all error-backprop  related paper reviews I have not yet found a case in which the authors did not admit that they used USUTS when I asked.  The second author of \cite{Graves16} claimed to have used PSUVS through a personal email to me but the first author who probably performed the experiments did not claim the same.  No authors of \cite{Graves16} responded to PubPeer questions towards \cite{Graves16}.

Weng 2020 \cite{WengFraudFaceBook2020,WengFraudYouTube2020} argued that the claims by some public speakers that such misleading errors have approached or even succeeded human performance \cite{Russakovsky15} are controversial, since there are no explicit competition rules that ban 
test sets to be used for Post-Selections. 

\subsection{Human PSUTS}
Instead of writing a search program in machine PSUTS, human PSUTS defined below typically involves less computational resources and programming demands. 

\begin{definition}[Human PSUTS]
After planning experiments or knowing what will be in the training set $T$ and test set $T'$, a human post-selects features in networks instead of using a machine to learn such features.
\end{definition}

Unfortunately, almost all methods in the symbolic school use human PSUTS because it is always the same human who plans for and design a micro-world and collect the test set $T'$.  The key to an acceptable 
test score lies in how much detail the human designer can plan for what is in the test sets and how much 
freedom s programmer has in hand picking features. 

Poggio et al. \cite{Serre07} and Fukushima et al. \cite{Fukushima83} explicitly admitted their use of human PSUTS.  Li Fei-Fei at al. \cite{Fei-Fei06} only vaguely admitted their use of human PSUTS by a vague term ``weakly supervised'' using an extension of formulation by Pietro Perona that is originally unsupervised.
Questions raised towards \cite{Fei-Fei06} on PubPeer were not answered by the authors.

\section{Why Error-Backprop Needs PSUTS}
\label{SE:WhyNeed}

This section discusses a global view, which is new as far as the author is aware, about why error-backprop  suffers from local minima even in the easier batch-learning mode. 

Since error-backprop does not perform well for incremental learning mode as we can see why also from the following discussion, we will concentrate on batch learning mode.   Namely, we let the  network ``see'' the entire training set $T$ for each network update. 

Let us first consider a well-known neuronal model that is applicable to many CNNs.  Suppose a post-synaptic neuron with activation $z_j$ is connected to its pre-synaptic neurons $y_i$, $i=1, 2, ... , n$, through synaptic weights $w_{ij}$, by the expression:
\begin{equation}
\phi (\sum_{i=1}^n w_{ij} y_i ) = z_j
\label{EQ:activation}
\end{equation}
where $\phi (y) = \frac{1}{1+e^{-y}}$ is the logistic function.   The gradient of $z_j$ with respect to 
weight vector $\w_j = (w_{1,j}, w_{2,j}, ... , w_{n,j}) $ is 
\[
\eta (y_1, y_2, ... , y_n) \triangleq \eta \y
\]
where $\eta$ is the partial derivative of $\phi (y)$.   Thus, according to gradient direction, the change of the weight vector $\w_j$ is along the direction of
pre-synaptic input vector $\y$.  If the error is negative, $z_j$ should increase.  Then the weight vector 
should be incremented by 
\begin{equation}
\w_j  \leftarrow \w_j + w_2 \y
\label{EQ:error-backprop}
\end{equation}
where $w_2$ is the learning rate.  We use the $w_2$ to relate better the optimal Hebbian learning, called LCA, used by DN in Section~\ref{SE:WhyNotNeed}.  At this point, the following theorem is in order.

\begin{theorem}[Lacks of error-backprop]
\label{TM:lack}
Error-backprop lacks (1) energy conservation, (2) an age-dependent learning rate, and (3) competition based
role-determination.
\end{theorem}
\begin{IEEEproof}
Proof of (1):  If pre-synaptic input vectors $\{\y\}$ are similar, multiple applications of Eq.~\eqref{EQ:error-backprop} add 
many terms of $\{w_2 \y\}$ into the weight vector $\w_j$ causing it to explode, which means a lack of energy conversation.   Proof of (2):  $w_2$ is typically tuned by an {\em ad hoc} way, such as a handpicked small value turned by a term called
momentum, instead of being automatically determined in Maximum Likelihood optimality (ML-optimality) by neuronal firing age to be discussed in Section~\ref{SE:WhyNotNeed}.   Proof of (3):  Suppose neuron $z_j$ is in a hidden area of the network hierarchy.   This neuron $z_j$ updates its pre-synaptic weight
using Eq.~\eqref{EQ:error-backprop} regardless $z_j$ is role-responsible or not for the current network error.   Likewise, looking upstream, there is also a lack of role-determination in the gradient-based update for pre-synaptic neurons $y_1, y_2, ... , y_n$, all of which must update their own weights using their own gradients.  Namely, there is no competition-based role-determination in error-backprop .
\end{IEEEproof}

\begin{figure}[tb]
	\centering
	\includegraphics[width=0.5\linewidth]{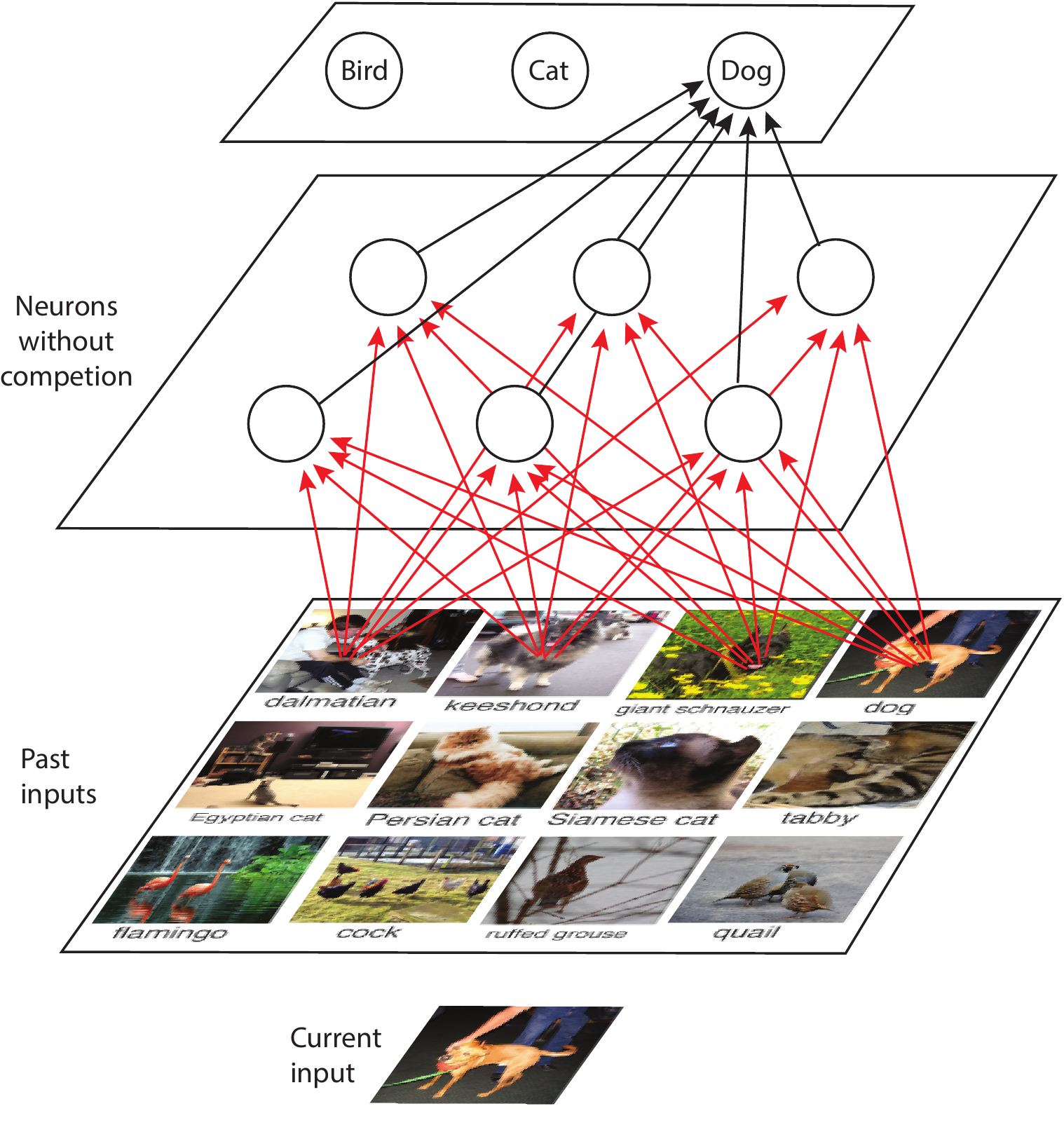}
	\caption{Lack of role-determination in hidden neurons due to a lack of competition.  The same ideas are true for a deeper hierarchy.  Color sample images courtesy of \cite{Russakovsky15}.
}
	\label{FG:LocalMinima-Complete-Connect}
\end{figure}

The meaning (3) of Theorem~\ref{TM:lack} are illustrated by Fig.~\ref{FG:LocalMinima-Complete-Connect}.
CNNs do not have a competition mechanism in any layers.  Complete connections initialized with random weights are provided for all consecutive areas (also called layers), from input area all the way to the output area.  
If the $z_j$ neuron is in the output motor area and each output neuron is assigned a single class label, 
the role of $z_j$ (``dog'' in the figure) is determined by human supervised label ``dog''.   However, let us assume instead that $z_j$ is in a hidden area, not responsible for 
the ``dog'' class.  $z_j$ still updates its input weights using the gradient.   Likewise, the pre-synaptic area $Y$, is characterized by its label ``neurons without competition''.  The hidden neurons in this area do not have a competition mechanism which would, like in LCA \cite{WengLCA09}, allow a small proportion of neurons to win the competition and fire so that they automatically take the roles that they happened to compete well.    This analysis leads us to the following theorem.

 \begin{theorem}[Random roles in error-backprop ]
 \label{TM:roles}
A set of random initial weights in a network assigns random roles to all hidden neurons, from which a 
local minimal point based on error-backprop learning inherits this particular random-role assignment.  Which neurons in each hidden area take a role does not matter, but how hidden neurons share a set of roles in each hidden area does matter in the final fitting error, validation error, and test error after error-backprop.
 \end{theorem}
 \begin{IEEEproof}
 Without loss of generality, suppose a maximum in the output neuron means a positive classification and weights take positive and negative values.  
 Then, a positive weight to an output neuron $z_j$ from a hidden neuron $y_i$ means an excitatory role of 
 $y_i$ to $z_i$ and a negative weight means an inhibitory role.   A zero weight means an irrelevant role.  
 The gradient vector computed in Eq.~\eqref{EQ:error-backprop} means such excitatory-inhibitory input patterns from pre-synaptic neurons are added through iterative error-backprop procedures.  Because of the complete connections and an identical neuronal type, where a hidden neuron is located in the Fig.~\ref{FG:LocalMinima-Complete-Connect} does not matter, but each input image must have a sufficient number of hidden neurons in every hidden area to excite for its signals to reach the corresponding output neuron.  The initial role assignment patterns in initial weights do matter for the final the fitting error rate, the validation error rate, and the test error rate, because gradient updates are local and inherited such initial roles.   
 \end{IEEEproof}
 
  \begin{theorem}[Percentage luck of error-backprop ]
 \label{TM:percentage}
Suppose a CNN has $l>1$ areas, $A_0, A_1, ... , A_l$, connected by a cascade or a variation thereof.  $A_0$ takes input frames $\{\x\in X\}$ and $A_l$ is the output area for classification.  Suppose an area has a total of $m$ hidden neurons that share a common receptive field $R$ in $A_0$.  Consider a given input frame $\x$.   Let the percentage of the $m$ hidden neurons that do not fire among all neurons in the same area with the same receptive field be denoted as $p(\x)$.  Then, the error-backprop  depends on the  average $\bar{p}=E_{\x\in X}\{p(\x)\}$ to be a reasonably small value, called {\em the percentage luck}. 
 \end{theorem}
\begin{IEEEproof} To guide the proof, we should mention that DNs use top-k competition so that each receptive field in each area has only $k$ neurons that fires, where $k$ is small, e.g., top-1, for each receptive field $R$.  Suppose a receptive field $R$ represents a neuronal column that has $n$ neurons.  A neuronal area at level $l$ is denoted as $A_l$. Every receptive field image $\x\in X=A_0$ is concrete by which we means that its neurons are only 
pixels $\{x\}$ of a concrete example of a class $C$ with $\bar{p}(x) = P(\mbox{$x$ fires}) \approx 50\%$ (e.g., 50\% back and 50\% white).  Each neuron $z$ in area $A_l$ is abstract by which we means that it fires means an abstract class $C$ that $\x$ belongs to, with 
$\bar{p}(z)$ being small corresponds to top-1 among $n$ neurons.   Then, it is necessary for the CNN to convert the most concrete representation of pixels in $A_0$ to more abstract representations in $A_l$, $l>0$ with a low $\bar{p}(z)$.  For example, in Fig.~\ref{FG:LocalMinima-Complete-Connect}, we have $l=2$ and there is no completion in the hidden area $A_1$.  Then error-backprop  depends on that each neuron in $A_2$ has only a relatively smaller percentage among $n=6$ neurons in $A_1$ that are positive, i.e., as the features of its particular class.   The requirement of being a small percentage is due to the need for other non-firing neurons to deal with many other patterns in the same receptive field.
\end{IEEEproof}

As we can expect, such a low percentage condition is rarely satisfied by a random weight vector.  The more random weight vectors one uses, the better chance to hit the luck.  
 
From Theorems~\ref{TM:lack} through \ref{TM:percentage} 
and their proofs, we can see that the luck of role assignment is a critical flaw of error-backprop , and so are the system parameters and the simple-minded regularization of the learning rate.  Because of these key reasons, PSUTS plays a critical role to 
select the luckiest network from many unlucky ones after error-backprop .  The more networks have trained by error-backprop , the more likely the luckiest one has a good role-assignment to start with. 

There has been no lack of papers that claim to justify error-backprop  does not over fit, e.g., variance based 
stochastic gradient decent \cite{Schaul13}, saddle-free deep network \cite{Dauphin14}, drop out \cite{Srivastava14}, implicit regularization during gradient flow \cite{Poggio20}.  They all address only local 
issues of neural networks trained by error-backprop  and did not mention Post-Selections.   The theory here addresses, the global role-assignment problem of random weights that no local mechanisms can deal with.   
This seems to be why PSUTS is necessary by error-backprop , but PSUTS is controversially fraudulent in terms of protocol---test sets are meant to test a reported system, not supposed to be used to decide which network to report from many.    

\section{How a DN Avoids Post-Selections}
\label{SE:WhyNotNeed}

Apparently, a brain does not use Post-Selection at al, whether UPSVS or PUUTS, because every human child must develop in a human environment to make his living.   He should not be covered up and not reported, regardless how well or bad he performs. 
Cresceptron in the 1990s \cite{WengIJCNN92,WengICCV93,WengCresIJCV97,WengPlagiarismFaceBook2020,WengPlagiarismYouTube2020} and later DN \cite{WengWhy11,WengIJCNN2020,WengIJHR2020,WengCAI-ICDL20} were inspired by the interactive mode that brains learn though lifetime.   In other words, Cresceptron and DN do not need Post-Selections. 
Furthermore, every DN must be ML-optimal given the same Three Learning Conditions. 
  
\subsection{New AI Metrics: Developmental Errors}
\label{SE:DevError}

In contrast to Post-Selections likely used by \cite{Werbos94,Fukushima80,Fei-Fei06,Serre07,Krizhevsky12,Karpathy14,LeCun15,Mnih15,Graves16,Moravcik17,McKinney20,Bellemare20} including probably AlphaGo \cite{Silver16}, AlphaGo Zero \cite{Silver17}, AlphaZero \cite{Silver18}, AlphaFold \cite{Senior20} and MuZero \cite{Schrittwieser20} and many others, we define and reported developmental errors that includes all errors occurred through lifetime of each learning network:
\begin{definition}[Developmental error]
A Developmental Network is denoted as $N=(X, Y, W_y, Z, W_z, A)$ with sensory area $X$, skull closed hidden area $Y$ and its weight space $W_y$, and motor area $Z$ and its weight space $W_z$, and the space of architecture parameters $A$, where $X$, $Y$, and $Z$ also denote the spaces of responses of $X$, $Y$ and $Z$ areas, respectively.  The space of architecture parameters $A$ includes all remaining parameters and memory of the network, other than neuronal weighs, such as ages of neurons (for learning rates), neuronal patterning parameters (location and receptive fields adapted by synaptic maintenance), neuronal types (for initial connection absences among areas), and neuronal growth rates (for speed of mitosis).  It runs through lifetime by sampling at discrete time indices as $N(t)$, $t=0, 1, 2, ... $ .   Start at inception $t=0$ with supervised sensory input $\x_0\in X(0)$, initial state $\z_0\in Z(0)$, randomly initialized weigh vector $\y_0 \in Y(0)$, initial architecture $\a_0 \in A(0)$.  At each time $t$, $t=1, 2, ... $, the network $N(t)$ recursively and incrementally updates:
\begin{equation}
(\x_t, \y_t, \z_t, \a_t) = f (\x_{t-1}, \y_{t-1}, \z_{t-1}, \a_{t-1}) 
\label{EQ:tupdate}
\end{equation}
where $f$ is the Developmental Program (DP) of $N$.  If $\z_t\in Z(t)$ is supervised by the teacher, the network complies and the error $e_t$ is recorded, but if the supervised motor vector has error, the error should be treated as teacher's.   Otherwise, the learner is not motor-supervised and $N(t)$ generates a motor vector $\z_t$ and is observed by the teacher and its vector difference from the desired $\z^*_t$ is recorded as error $e_t$.  The lifetime average error for each motor concept or component, from time $0$ up to time $t$ is defined as 
\begin{equation}
\bar{e}(t) \triangleq \frac{1}{t} \sum_{i=0}^t e_i,
\end{equation}
which is computed incrementally in terms of average developmental error $\bar{e}(t)$: 
\begin{equation}
\bar{e}(t) = \frac{t-1}{t} \bar{e}(t-1)  +  \frac{1}{t} e_t.
\end{equation}
\end{definition}

Namely, all errors across a lifetime, at every time instance, are caught by the developmental error.  In order to reach a small error, a low final error rate that a batch learning method tries to reach is not sufficient.  Instead, the network must learn as fast as possible and avoid errors as much as possible at every time instance $t$.  This is indeed important since earlier performance will shape later learning.

An optimal network that gives the lowest possible developmental error, among all possible networks under the same Three Learning Conditions, must be optimal at every time instance $t$ throughout its life.  DN is one such network.   Post-Selections are useless among neural networks that give the smallest developmental error under the same Three Learning Conditions, because the maximum-likelihood optimality  should give equivalent networks of the same developmental error.   

However, in practice, the learning experience in the Three Learning Conditions is unlikely the same among different networks, because each physical robot that runs a network at least occupies distinct physical locations in the real world.  For example, if two physical robot in the same family fight for a toy, the winner gains a winner experience and the loser may acquire a loser mentality.   In other words, even if the parents of two boys are not biased toward any boys, the competition among the boys results in different learning experiences.

The developmental error is important.  If a competition is based on developmental errors (such as during AIML Contests \cite{WengAAAIFS18}), the winner is unlikely be one that 
uses a brute force method but has an excessive amount of computational resources and manpower.
ImageNet competitions \cite{Russakovsky15} are flowed also in this sense. 

Although not formally defined as developmental errors, Cresceptron \cite{WengCresIJCV97} and Developmental Networks \cite{WengIJIS15,WengPatentDN-2,Knoll-ICDL21} reported developmental errors.

Namely, the developmental 
error, unless stated otherwise for a particular time period, is the average lifetime error from inception.  
To report more detailed information about the process of developmental errors $\{\e_t \;|\; t\ge 0\}$, statistics other than the mean (average) can be utilized, such as the minimum, 
25\%, 50\% (median), 75\%, the maximum, and the standard deviation. 

For more a specific time period, such as the period from age
$t_1$ to age $t_2$, the average error is denoted as $\bar{e}[t_1:t_2]$.  Therefore, $\bar{e}(t)$ is a short notation for $\bar{e}[0:t]$.

Because Cresceptron and DN have a dynamic number of neurons up to a system memory limit, each new context
\begin{equation}
\c_t  \triangleq (\x_t, \y_t, \z_t)
\end{equation}
may be significantly different from the nearest matched learned weight vectors of all hidden neurons.   
If that happens and there are still new hidden neuron that have not fired, a free-state neuron that happens to be the best match is spawned that perfectly memorizes this new context regardless its randomly initialized
weights.  When all the free neurons have fired at least once, the DN will update the top-$k$ matched neurons optimally in the sense of maximum likelihood (ML), as proven for DN-1 by \cite{WengIJIS15} 
and for DN-2 by \cite{WengPatentDN-2}, as we will discuss below.

Note that a developmental system has two input areas from the environment, sensory $X$ and motor $Z$.   That is, motor $Z$ is supervisable by the world (including teachers) but not often.   Since there is hardly any sensory input $\x\in X$ that exactly duplicates at two different time indices, 
almost all sensory inputs from $X$ are sensory-disjoint.   During motor-supervised learning, if the 
teacher supervises its motor area $Z$ and the learner complies.   Since a teacher can make an error, 
the motor-error that the teacher made is also recorded as the developmental error of the motor of the learner but due to the teacher.  

\subsection{Neuronal Competitions}

As discussed above, error-backprop  learning is without neuronal competitions.   The main purpose of competition is
to automatically assign roles to hidden neurons.  Below, we consider two kinds of Convolution Neural Networks (CNNs), sensory networks and sensorimotor networks.  A sensory network is feedforward, from sensor to motor, in computation flow and therefore is simpler and easier to understand.  A sensorimotor network takes both sensor and motor as inputs and is highly recurrent and therefore more powerful. 

\subsubsection{Sensory networks}

\begin{figure*}[tb]
	\centering
	\includegraphics[width=1.0\linewidth]{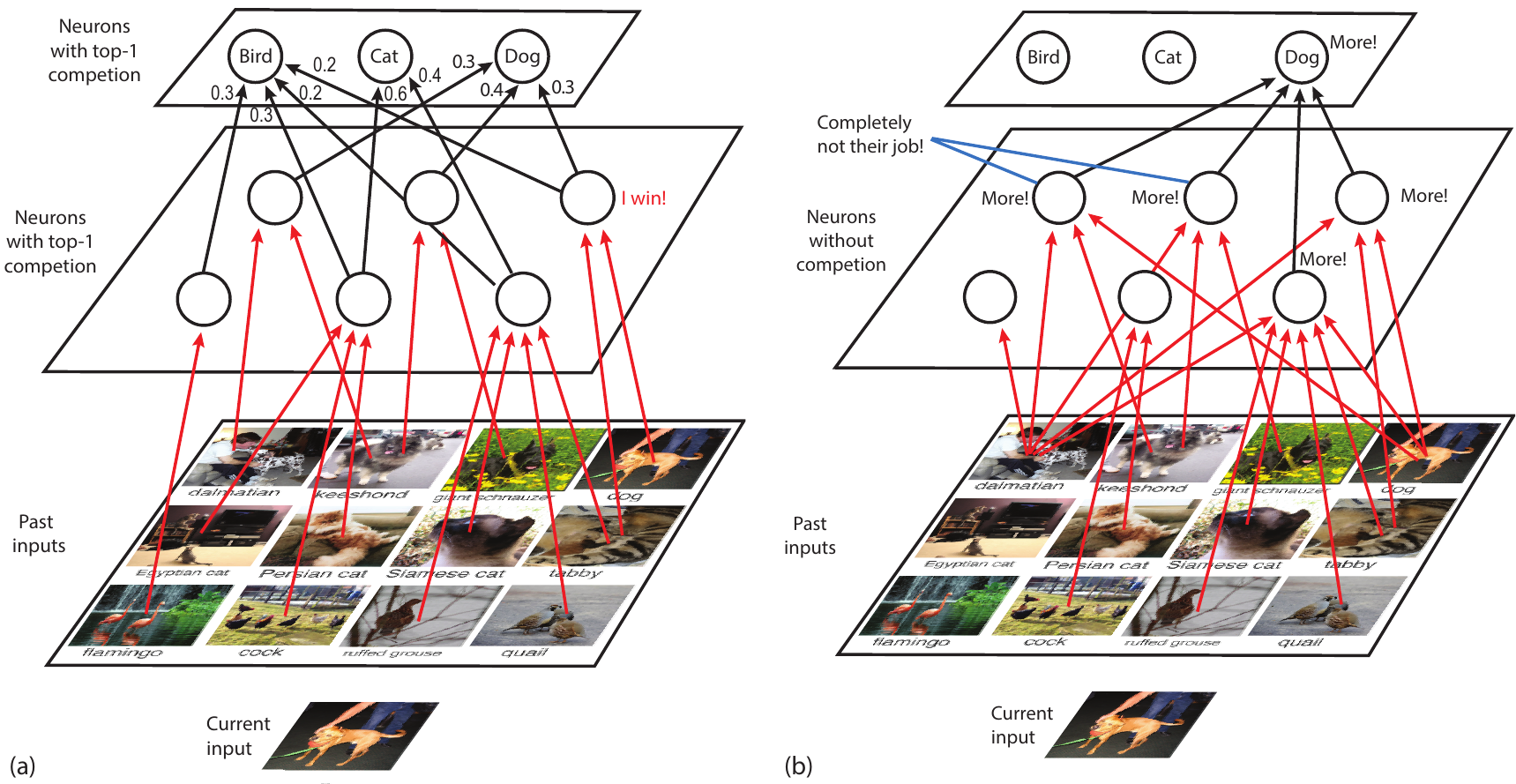}
	\caption{How competition automatically assigns roles among hidden neurons without a central controller: The case for automatically construct a mapping $f: X \mapsto L$. 
	(a) The number of samples in $X$ is larger than the number of hidden neurons such that each hidden neuron must win-and-fire for multiple inputs.
	(b) Error-backprop from the ``dog'' motor neuron asks some hidden neurons to help but the current input feature is not their job.  Thus, error-backprop  messes up with the role assignment guessed by the random initial weights.  The same ideas are true for a deeper hierarchy.
	Color sample images courtesy of \cite{Russakovsky15}.
}
	\label{FG:LocalMinima-2-subfigs}
\end{figure*}

Let us first consider the case of feed-forward networks as illustrated in Fig.~\ref{FG:LocalMinima-2-subfigs}.  Fig.~\ref{FG:LocalMinima-2-subfigs}(a) shows a situation where the number of samples in $X$ is larger than the number of hidden neurons, which is typical.  Otherwise, if there are sufficient hidden 
neurons, each hidden neuron can simply memorize a single sample $\x\in X$.  

This means that the total number of hidden neurons must be shared through incremental learning, where each sample image-label pair $(\x, l)\in X\times L$ arrives incrementally through time, $t=0, 1, 2, ... $.  This is the case with Cresceptron (and some other networks) which conducts incremental learning by dealing with image-label pairs one at a time and update the network incrementally.  

Every layer in Cresceptron consists of a image-feature kernel, which is very different from those in DN where each hidden neuron represents a sensorimotor feature to be discussed later.   
By image-feature, we mean that each hidden neuron is centered at an image pixel.  Competitions take place within the column for a receptive field centered at each pixel at the resolution of the layer.  The resolution reduces from lower layer to higher layer through was called resolution reduction (also called drop-out). 

The competition in incremental learning is represented by incrementally assigning a new neuronal plane 
(convolution plane) where the new  kernel memorizes the new input pattern if the best matched neuron in a column does not match sufficiently well.   Suppose images $\x \in X$ arrives sequentially, 
the top-1 competition in the hidden layer in Fig.~\ref{FG:LocalMinima-2-subfigs}(a) enables each hidden neuron to respond to multiple features, indicated by the typically multiple upward arrows, one from each image, pointing to a hidden neuron.   This amounts to incremental clustering based on top-$k$ competition. The weight vector of each hidden $Y$ neuron corresponds to a cluster
in the $X$ space.   In Fig.~\ref{FG:LocalMinima-2-subfigs}(a), $k=1$ for top-$k$ competition in $Y$.

Likewise, suppose top-1 competition in the next higher layer, $Y$, namely each time only one $Y$ neuron fires at 1 and all other $Y$ neurons do not fire, resulting the connection patterns from the second layer $Y$ to the next higher layer $Z$.  In the output layer $Z$, top-1 competition takes place but a human teacher can supervise the pattern. 

The Candid Covariance-free Incremental (CCI) Lobe Component Analysis (LCA) in Weng 2009 \cite{WengLCA09} proved that such automatic assignment of roles through competition results in 
a dually optimal neuronal layer, optimal spatially and optimal temporally.   Optimal spatially means the CCI LCA incrementally computes the first principal component features of the receptive field.  Optimal temporally means that the principal component vector has the least expected distance to its target---the optimal estimator in the sense of minimum variance to the true LCA vector. 

Intuitively, regardless what random weights each hidden neuron starts with, as soon as it wins to fire for the first time, its firing age $a=1$.  Its random weight vector is multiplied by the zero retention rate $w_1= 1-1/a=0$ and this learning rate $w_2= 1/a=1$ so that the new weight vector becomes the first input $r\x$ with $r=1$ for the firing winner.
\begin{equation}
\v \leftarrow(1 -\frac{1}{a}) \v + \frac{1}{a} r\x.
\label{EQ:winner}
\end{equation}
It has been proven that the above expression incrementally computes the first principal component as $\v$. 
The learning rate $w_2=\frac{1}{a}$ is the optimal and age-dependent learning rate.  CCI LCA is a framework for dually optimal Hebbian learning.  The property ``candid'' corresponds to the property that sum of 
the learning rate $w_2 =  \frac{1}{a}$ and the retention rate $w_1= 1 -\frac{1}{a}$ is always 1 to keep the 
``energy'' of response $r$ weighted input $\x$ unchanged (e.g., not to explode or vanish).   This dually optimality resolves the three problems in Theorem~\ref{TM:lack}.

Fig.~\ref{FG:LocalMinima-2-subfigs}(b) shows how the three neurons in the $Z$ area updates their weights so that the 
weight from the second area to the third area become the probability of firing, conditioned on the firing of the 
post-synaptic neuron in area $Z$ (Dog, Cat, Bird, etc.).  The CCI LAC guarantees that the sum of weights for each 
$Z$ neuron sum to 1.  This automatic role assignment optimally solves the random role problem of 
error-backprop  in Theorem~\ref{TM:roles}.

However, optimal network for incrementally constructing a mapping $f: X\mapsto L$ is too restricted, since
$f: X\mapsto L$ is only what brains can do, but not all brains can do.  For the latter, we must address sensorimotor networks. 

\subsubsection{Sensorimotor networks}
The main reason that Marvin Minsky \cite{Minsky91} complained that neural network is scruffy was because 
conventional neural networks lacked not only the optimality described above for sensory networks, but also lacked the Emergent Universal Turing Machines (EUTM) that is ML-optimal we now discuss below. 

First, each neuron in the brain not only corresponds to a sensory feature as illustrated in Fig.~\ref{FG:LocalMinima-2-subfigs}, but also a sensorimotor feature.   By sensorimotor feature, we mean 
that the firing of each hidden neuron in Fig.~\ref{FG:LocalMinima-2-subfigs} is determined not just by 
the current image $\sigma$ represented by a sensory vector $\x \in X$, but also the state $q$ represented by a motor vector $\z\in Z$.   It is well known that a biological brain contains not only bottom-up inputs from $\x \in X$ but also top-down inputs from $\z \in Z$.  In summary, each hidden neuron represents a sensorimotor feature in a complex brain-like network. 

\subsection{FA as sensorimotor mapping}
This sensorimotor feature is easier to understand if we use the conventional symbols for (symbolic) automata.  Let us borrow the idea of Finite Automaton (FA).   In an FA, transitions are represented by function $\delta: Q \times \Sigma \mapsto Q$, where $\Sigma$ is the set of input symbols and $Q$ the set of states.  Each transition is represented by
\[
(q, \sigma) \stackrel{f}{\longrightarrow} q'
\]

\begin{figure}[h]
	\centering
	\includegraphics[width=0.4\linewidth]{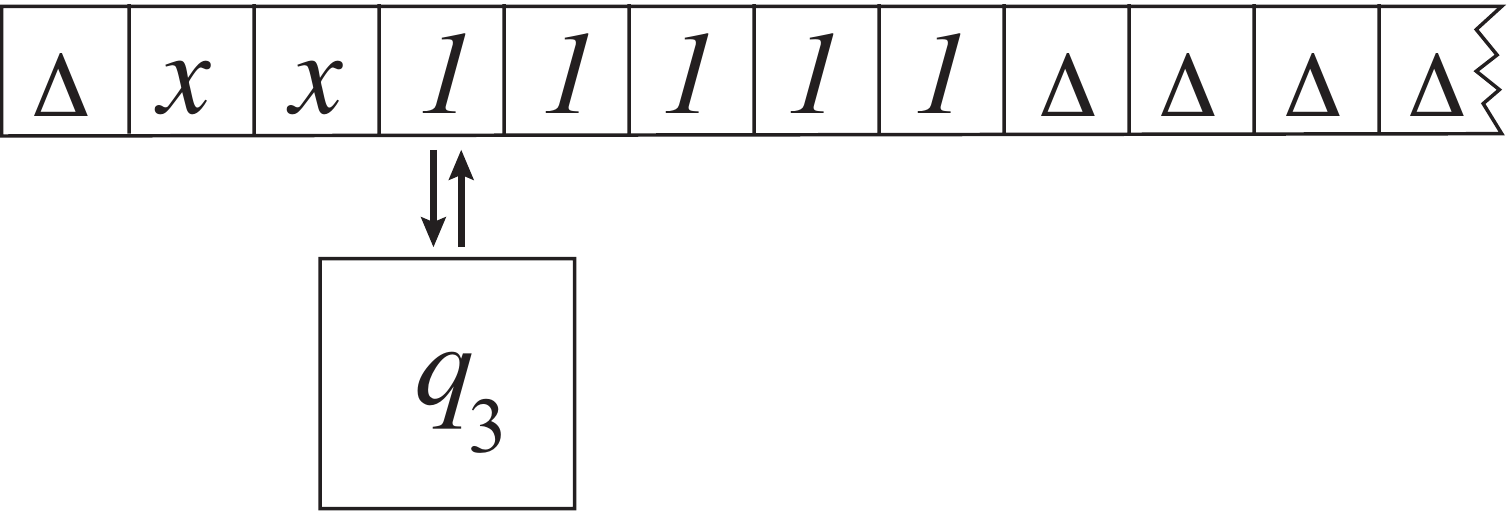}
	\caption{
	A Turing machine has a tape, a read-write head, and a transition function with a current state.}
	\label{FG:Turing-Machine-function}
\end{figure}

\paragraph{AFA as a control of any Turing machine}
Weng 2015 \cite{WengIJIS15} extended the definition the FA so that it outputs its state so the resulting FA becomes an Agent FA (AFA).   Further, Weng 2015 \cite{WengIJIS15} extended the action $q$ to the 
machinery of Turing machine (see Fig.~\ref{FG:Turing-Machine-function}) so that action $q$ includes output symbol to the Turing tape and the head motion of the read-write head of a Turing machine.   With this extension, Weng 2015 \cite{WengIJIS15} proved that the control of any Turing machine is an AFA, a surprising result.   

Here $q\in Q$ is the top-down motor input to a sensorimotor feature neuron;  $\sigma$ is the 
bottom-up sensory input to the same neuron.   If $\delta$ has $n$ transitions, $n$ hidden neurons in the $Y$ area are sufficient to memorize all the transitions that is observed sequentially, one transition at a time. 

We should not use symbols like $\sigma$ and $q$, but instead sensory vectors $\x\in X$ and motor vectors $\z \in Z$ that are emergent as discussed above.   At discrete time $t=0, 1, 2, ... $, we use the hidden neurons in the $Y$ area to incrementally learn the transitions:
\begin{equation}
\begin{bmatrix}
Z(0)\\
Y(0)\\
X(0)
\end{bmatrix}
\rightarrow
\begin{bmatrix}
Z(1)\\
Y(1)\\
X(1)
\end{bmatrix}
\rightarrow
\begin{bmatrix}
Z(2)\\
Y(2)\\
X(2)
\end{bmatrix}
\rightarrow
...
\label{EQ:time}
\end{equation}
where $\rightarrow$ means neurons on the right use the input neurons on the right and compete to fire as explained below without iterations.  Namely, by unfolding time, the spatially recurrent DN becomes non-recurrent in a time-unfolded and time-sampled DN.   With LCA update, \cite{WengIJIS15} proved that such a DN is ML-optimal and has a constant complexity for each update $O(1)$ with a large constant, suited for real-time computation with a large memory and many neurons.   

\subsection{DN as a ML-Optimal of Emergent Universal Super-Turing Machine}
The traditional Turing Machine (TM) is a human handcrafted machine, as illustrated in Fig.~\ref{FG:Turing-Machine-function}.   

A Universal TM (UTM) s still a TM, but its
tape contains two parts, a user supplied program and the data that the program is applied to.   The transition function of the UTM is designed to simulate any program encoded in the form of transition of a TM and
to apply the program on the data on the tape and finally to place the output of the program on the data onto the 
tape.  

A UTM is a model of the current general-purpose computers because the user can write any program on any set of appropriate data for the UTM to carry out.   Because a DN is an ML-optimal emergent 
FA, Weng 2015 \cite{WengIJIS15} extended a symbolic Turing machine to a super Turing machine
by (1) extending the tape to the real world, (2) the input symbols to vectors from sensors, (3) the output
symbols to vector output from effectors, and (3) the head motion to any action from the agent.  
Thus, DN ML-optimally learns any TM, including UTM, directly from the physical world.   The programs on the tape are learned by the Super UTM incrementally from the real world across its lifetime!

\subsection{DN as a ML-Optimal Learning Engine for APFGP and Conscious Learning}
Because DN is an ML-optimal learning engine for any TM, including UTM, DN ML-optimally learns 
any UTM from the physical world, conditioned on those in Definition~\ref{DF:Conditions}.  This means that a DN ML-optimally learns to Autonomous Programming for General Purposes (APFGP) \cite{WengIJHR2020,WengJCS2020}.   Based on the capability of APFGP,
Weng 2020 argued that APFGP is a characterization of conscious machines \cite{WengCAI-ICDL20} that boots its skill of
consciousness through conscious learning---being (partially) conscious while learning across lifetime.   Hopefully, APFGP is a clearer and more precise characterization for conscious machines and animals, assuming that we allow a conscious machine to develop its degree of consciousness from infancy. 

In the following, we list the DN algorithm so that we can understand APFGP is not a vague idea and how APFGP by DN avoids Post-Selections.

\subsection{DN-2 Algorithm}

Let us go through the DN-2 algorithm here so that we can see that DN is fully detail in computer implementation.  

DN-2 is the latest general-purpose learning engine in the DN family. In DN-1, the allocation of neurons in each subarea of the hidden $Y$ area is handcrafted by the designer. In DN-2, several biology-inspired mechanisms are added to automatically allocate neuronal resources and generate a dynamic and fluid hierarchy of internal representations during learning, relieving the human designer from handcrafting a concept hierarchy, beyond the 
rigid hierarchy in deep learning \cite{WengIJCNN92,WengICCV93,WengCresIJCV97,WengPlagiarismFaceBook2020,WengPlagiarismYouTube2020,Fei-Fei06,Serre07,Krizhevsky12,Karpathy14,LeCun15,Mnih15,Graves16,Moravcik17,McKinney20,Bellemare20}.   Namely, a DN-2 starts with simple internal representations which gradually grow to be 
rich and deep supported by early representations as a ``brain stem'', but it is still ML-optimal conditioned on those in 
Definition~\ref{DF:Conditions}.

Areas from low to high: $X$: sensory; $Y$ hidden (internal); $Z$: motor.  From low to high: bottom-up.  From high to low: top-down.  From one area to the same area: lateral.   $X$ does not link with $Z$ directly.\\

Input areas: $X$ and $Z$;
Output areas: $Z$; Hidden area: $Y$, fully closed from $t=0$.

\begin{enumerate}
\item At time $t=0$, inception. Initialize the $X$, $Y$ and $Z$ areas.  $\x \in X$ takes the first image. 
Set every $Y$ neuron with random weights, zero firing age, and zero response $\y(0)$. Set the total number of $Y$ neurons to be $n_Y$.  A boundary $c_Y$ indicates the number of active neurons ($c_Y \le n_Y$).  Set  the $Z$ area and its memory part $M_Z$ similarly, but all concept zones take none vectors if the learner has  no prenatally learned inborn ``reflexes''. 
\item For time $t=1, 2, ...$, repeat the following steps forever (executing steps \ref{step_a}, \ref{step_b} in parallel, before step \ref{step_c}):
\begin{enumerate}
\item \label{step_a} All $Y$ neurons compute in parallel:
\begin{equation}
\label{y_area}
(\y', M'_Y) = f_Y(\c_Y, M_Y)
 \end{equation}
where context $\c_Y=(\x,\y,\z)$, $M_A$ denotes the memory of area $A$ including weights and neuronal firing ages, and $f_Y$ is the $Y$ area function using LCA \cite{WengNAI2ed2019, WengLCA09}. If the best active $Y$ neurons do not match the input vector well, area $Y$ transfers new neurons to active and increment the boundary $c_Y$.

\item  \label{step_b} Supervise $\z'$ if the teacher likes. Otherwise, $Z$ neurons compute the response vector $\z$  and update memory $M'_Z$ in parallel:
\begin{equation}
\label{z_area}
 (\z', M'_Z)=f_Z(\c_Y, M_Z)
\end{equation}
where $f_Z$ is the $Z$ area function using LCA \cite{WengNAI2ed2019,WengLCA09} and $\c_Z=(\y,\z)$. 
\item \label{step_c} Replace asynchronously: $(\y, M_Y, \z, M_Z) \leftarrow (\y', M'_Y, \z' , M'_Z)$.  Supervise input $\x$.
\end{enumerate}
\end{enumerate}

The area function $f_Y$ in Eq.\eqref{y_area} and area function $f_Z$ in Eq.\eqref{z_area} include two parts: (1) The computation of response vectors $\y'$ and $\z'$, respectively;
(2) The maintenance of memory $M'_Y$ and $M'_Z$ for $Y$ area and $Z$ area, respectively. 

The ML-optimality of DN-1 and DN-2 is rooted in the optimality of LCA and extends to the entire network and entire lifetime.

\subsection{Methods for Recursive Optimization}
\label{SE:methods}

Given the Three Learning Conditions, at each time $t$, $t=1, 2, ... $, a DN incrementally computes the ML-estimator of its parameters at each time $t$ that minimizes the developmental error without
doing any iterations.  

Let us first review the maximum likelihood estimator for a batch data.   Let $\x$ be 
the observed data and $f_\theta (\x, \z)$ is the probability density function that depends on a vector $\theta$ of parameters, there $\theta(t) = (\w_y, \w_z, \a)$ where some parameters of the architecture parameter vector $\a$ are
hand-initialized such as the receptive fields.  The maximum estimator for $\theta$ corresponds to the $\theta$ that maximizes the probability density. Regardless $\z$ is imposed, $\z$ is part of the parameters to be computed in a closed-form as a self-generated version:
\begin{equation}
(\theta^* , \y^*, \z^*)= \argmax_{(\theta, \y, \z)} f_\theta (\x, \z).
\label{EQ:ML-x}
\end{equation}

Since the above lifetime estimator is incremental, at each time $t$, the previous state $\z_{t-1}^*$ is self-generated or supervised, and the observation is $\x_{t-1}$.  The incremental ML-estimator for $\theta^*_{t} $ is computed in a closed-form by the incremental version of Eq.~\eqref{EQ:ML-x} where $f$ uses context $\c_{t-1}=(\x_{t-1}, \y_{t-1}, \z_{t-1})$:
\begin{equation}
(\theta^*_t , \y^*_{t}, \z^*_{t}) = \argmax_{(\theta_{t}, \y_{t}, \z_{t})} f _{\theta_{t}}(\x_{t-1}, \y^*_{t-1}, \z^*_{t-1}).
\label{EQ:ML-xz-inc}
\end{equation}
The DN computes the above expression for each time $t$ in a closed form without conducting any iterations \cite{WengIJIS15,WengPatentDN-2}.

How about initial weights?  Inside $\theta$, the weights of the DN are initialized randomly at $t=0$.
There are $k+1$ initial neurons in the $Y$ area, and $V = \{ \dot \v_i  \;|\; i=1, 2, ... , k+1\}$ is the current
synaptic vectors in $Y$.  Whenever the network takes an input $\p$, compute the
pre-responses in $Y$.  If the top-$1$ winner in $Y$ has a pre-response lower than almost perfect match $m(t)$ discussed below, activate a free neuron to fire.  Eq.~\eqref{EQ:winner} showed that the initial weights of this free neuron is multiplied by a zero and therefore do not affect its updated weights.  

Weng~\cite{WengIJIS15} proved that DN-1 computes the ML-estimator of all observations from the sensory space $X$ and motor space $Z$ using a large constant
time complexity for each time $t$.  Although DN learns incrementally, such a DN is error-free for learning any complex Turing machines, including any universal Turing machines.  
Weng~\cite{WengPatentDN-2} did the same for DN-2. 

\subsection{How DN Avoids Post-Selections but is Further ML-Optimal}

Since weights are initialized randomly, how does a DN result in an equivalent network regardless the random seed?
There are $k+1$ initial neurons in the $Y$ area, and $V = \{ \dot \v_i  \;|\; i=1, 2, ... , k+1\}$ is the current
synaptic vectors in $Y$.  Whenever the network takes an input $\p$, every $Y$ neuron computes the
pre-response.  If the top-$1$ winner in $Y$ has a pre-response lower than almost perfect match $m(t)$, activate a free neuron to fire. The almost perfect match $m(t)$ is defined as follows:
\begin{equation}
m(t) =  (1 - \delta)(1-e^{-t/t_1})
\label{EQ:match}
\end{equation}
where $\delta$ is the bound of machine round-off errors, and $t_1$ the childhood length.

Using a mathematical induction procedure, Weng~\cite{WengIJIS15} proved that DN-1 computes the ML-estimator of all observations from $(\x, \y, \z)$ using a large constant
time complexity for each time $t$.  Weng et al. 2018 \cite{WengPatentDN-2} proved the ML-optimality for  DN-2.   Since the number of transition of any Turing machine is finite, when the DN learns a Turing machine, a finite number of hidden $Y$ neurons is sufficient for the DN to incrementally memorize exactly all the transitions observed from the Turing machine.   In other words, although DN learns incrementally, such a DN is error-free for learning any complex Turing machines, including any universal Turing machines.   

If the DN runs in the real world, any finite size DN is not error-free soon after inception since the number of  observations from the real world is virtually unbounded, although each life is time bounded.   Namely, the
amount of data from the real world is so large that any practically large DN will eventually run out of
free neurons that have not fired yet.   From that point on, the DN is no longer guaranteed to be error-free, although could be sometimes error-free, but is still ML-optimal inside the skull conditioned on those in Definition~\ref{DF:Conditions}.   In other words, in the sense of ML, the DN is free of local minima inside the skull.  That is why only one DN is sufficient for each life and the DN avoids Post-Selections.   However, because the three conditions in Definition~\ref{DF:Conditions}, a DN is not be optimal unconditionally either.   For example, a better designed teaching schedule or a more appropriate physical environment may enable a DN to learn and discover rules faster and better. 

\subsection{Comparison with HMM}

It is important to compare the traditional Hidden Markov Model (HMM) \cite{Rabiner89a} with the ML-optimal DN.  (1) The former does not have any internal representations other than the symbol based
probabilities; the latter self-generates internal representations to generalize based on internal-representation based probabilities (e.g., weights).  (2) The former uses batch learning but the latter uses incremental learning.  (3) States in the former are symbolic, static, only partially observable for HMM, and not teachable but those in the latter are emergent, observable and directly teachable if the teacher like.  (4) The former requires a batch clustering method (e.g., k-mean clustering) to initialize a static set of symbolic states, but the states/actions in the latter are incrementally taught or autonomously generated and tried.   (5) Clusters of states in the former are not supported by a statistical optimality and the probability is only for state estimates but those in the latter are ML-optimal throughout the lifetime of learning, not in states/actions that the learner must produce and do not have a freedom for, but for the internal representations that the learner does have a high degree of freedom for.  (6) Due to the need to compute internal representations, the 
amount of computations in the latter is often higher than the former but the computational complexity is linear in time with a large constant (the number of weights of all available neurons). 

\section{Experiments}
\label{SE:Exp}

\subsection{Vision, Audition and Natural Languages}

The recent experimental results of DN work here include (1) vision that includes simultaneous recognition and detection and vision-guided navigation on MSU campus walkways \cite{Zheng19}, (2) audition to learn phonemes with a simulated cochlea and the corresponding behaviors \cite{WuIJCNN20}, (3) acquisition of English and French in an interactive 
bilingual environment \cite{Carstro-GarciaIJCNN19}, and (4) exploration in a simulated maze environment with autonomous learning
for vision, path cost, planning, and selection of the least-cost plan, where all such  
emergent actions are either {\em covert} (thoughts) or {\em overt} (acts) \cite{WuThink21}.   The same network was used to learn these four very different
tasks and task environments while each task embeds the ML-optimality of the network, under the Three Learning Conditions. 

\subsection{Error-Backprop vs. ML-Optimal DN}

\begin{figure}[tb]
	\centering
	\includegraphics[width=1.0\linewidth]{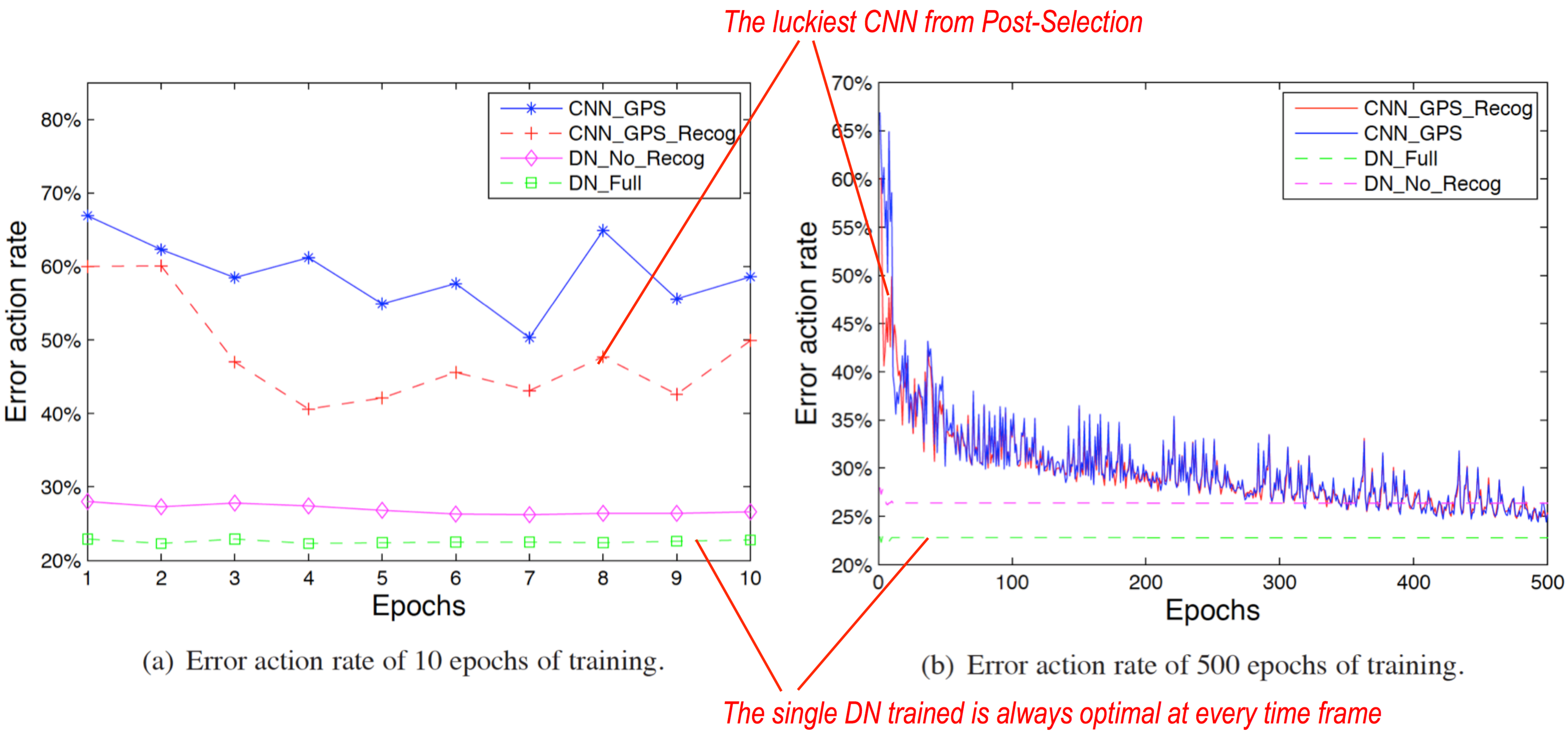}
	\caption{Comparison of error between the luckiest CNN trained by batch error-backprop and a 
	DN across different epochs through the training data.  ``Recog'' and ``Full'' means teach where-what rules.  Otherwise, data-fitting only.  Adapted from \cite{ZhengCVVT16}.
}
	\label{FG:CNN-vs-DN}
\end{figure}

To show the effects of the absence of ML-optimality in CNN vs. the ML-optimality of DN, Fig.~\ref{FG:CNN-vs-DN} shows the errors of the luckiest Convolutional Neural Network (CNN) trained by a batch error-backprop method and the errors of a DN trained incrementally.  As we understand, batch learning should not be compared with an incremental learning method, because it is not a comparison on an equal footing.  However, Fig.~\ref{FG:CNN-vs-DN} shows that DN does a harder (incremental) work drastically better than CNN does an easier (batch) work.  The task is real-world vision-guided navigation on the campus of Michigan State University.
 Because the DN is optimal in maximum-likelihood, it reaches the minimum error as soon as it has gone through the data set $T$ once (one epoch).  Later epochs correspond to reviews of the same data set $T$.  According to the maximum-likelihood principle, the optimal estimate of the neuronal weights should not change but the ages of the neurons continue to advance.  In contract, the luckiest error-backprop trained CNN chosen from several random seeds need many epochs to reduce its errors and only very slowly.  At the end of 500th epoch, the error of the luckiest CNN trained by error-backprop  is still considerably higher than the full DN.  Furthermore, as show in Fig.~\ref{FG:CNN-vs-DN}, teaching invariant concepts, i.e., abstraction in Theorem~\ref{TM:abs}, are use for reducing the optimal errors.  For more detail, the reader is referred to \cite{ZhengCVVT16}.

\subsection{AIML Contests}

In the AIML Contest 2016, all teams are required to use a single learning engine to learn three sensory modalities, vision, audition, and
bilingual natural languages acquisition, while the engine learns in ``lifetime''.  Although all the teams are free to choose any existing learning engines such as DN, tensor-flow or other engines, all the teams
chose DN engine (open source).  The supplied simulated sequential data (yes, subject to the ``big data'' flaw) are as follows.  When we let the $\x$ be the image at each time instance and $\z$ be the pattern of landmark location-and-type and action of navigation, the DN became a vision-guided navigation machine.
When we let $\x$ be the frame of firing pattern of hair cells in cochlea at each time instance and $\z$ be
the dense states/contexts and the sparse type of sounds, the DN became a auditory-recognizer machine.
When we let $\x$ be a time frame of vector of word (either English or French) and $\z$ be the language kind (neutral, English, and French) and meaning of each sentence context, the DN became a bilingual language learner and recognizer.   Thus, the AIML Contest appeared to be the first contest that independently demonstrated task-nonspecificity and modality-nonspecificity by independent laboratories with the contest teams.  All the teams are evaluated under the same Three Learning Conditions, e.g., the number of neurons in the engine must not exceed the same given bound.   The Contest used the developmental error like the one defined here averaged over all the three contest tasks and across the three lifetimes.  The {\em developmental error}  ranked all the submitted contest entries and required all networks not to exceed the specified maximum number of neurons for each task so that the competition does not unfairly favor those teams that have more computational resources at their disposal but not necessarily that their methods are more superior.   All the teams have a high degree of freedom to modify the learning engine and to modify the supplied motor actions on the given data set, such as generating attentive actions on the given training set to train invariant concepts (e.g., where and what concepts) which modifies the default training experience supplied by the AIML Contest organizers but was still based on the same supplied data set.   The prudent design of the AIML contests was meant to avoid the corresponding problems in ImageNet Contests \cite{Russakovsky15} and many other contests.

\subsection{GENISAMA Applications}

GENISAMA LLC, a startup that the author created, has produced a series of real-time machine learning products, as human-wearable robots.  They are the first products ever existed as APFGP robots. 
Hopefully, as a APFGP platform, this new kind of human-wearable robots will be useful for practitioners to
produce various kinds of intelligent auto-programed software.  The author predicts that such a new kind of AI systems will 
considerably alleviate the high brittleness of traditional AI software and traditional robot software in open and natural world.    

Hopefully, future DN-driven robots will learn consciously and autonomously discover in the real world for Turing machine based general purposes, with relative infrequent interactions from humans similar to what parents do to their children and human teachers teach their students in classrooms.   The experiments and competitions described here are for this grand goals but have not reached this experimental goal yet.

\section{Conclusions}
\label{SE:conclusions}

We used intuitive terms but formal ways to discuss Post-Selections.   Public and media have gained an impression that deep learning has approached or even ``sometimes exceeded'' human level performance on certain tasks.  For example, the image classification errors from a static image set were compared with those of humans \cite[A2, p242]{Russakovsky15}) and the work is laudable.  However, this paper raises Post-Selections, which seem to question such claims since a real human does not have
the luxury of Post-Selections.  The author hopes that the exposure of Post-Selections is beneficial to AI credibility and the future healthy development of AI, especially with the concepts of developmental errors and the framework of ML-optimal lifetime learning for invariant concepts under the Three Learning Conditions.   Some researchers have raised that it seems that those who wan a competition were those who have more computational resources and manpower at their disposal.   The new developmental error metrics under the Three Learning Conditions hopefully encourages 
future AI competitions to compare methods under the same Three Learning Conditions.  Considering DN as a much-simplified model for a biological machine, it seems not baseless to guess that 
each biological brain is probably ML-optimal (of course in a much richer sense) across lifetime,
e.g., due to the pressure to compete at every age.   The Three Learning Conditions explicitly include other factors that greatly affect machine learning performances such as learning framework (e.g., task-nonspecificity, incremental learning, the robot bodies), learning 
experiences and computational resources.   The analysis that any ``big data'' sets are nonscalable does not
mean that we should not create, use and share data sets.   Instead, we need to pay attention to the fundamental limitations of any static data sets, regardless how large their apparent sizes are.

\bibliographystyle{naturemag}
\bibliography{shoslifref}

\begin{thebibliography}{10}
\expandafter\ifx\csname url\endcsname\relax
  \def\url#1{\texttt{#1}}\fi
\expandafter\ifx\csname urlprefix\endcsname\relax\def\urlprefix{URL }\fi
\providecommand{\bibinfo}[2]{#2}
\providecommand{\eprint}[2][]{\url{#2}}

\bibitem{Montfort05}
\bibinfo{author}{Montfort, N.}
\newblock \emph{\bibinfo{title}{Twisty Little Passages: An Approach to
  Interactive Fiction}} (\bibinfo{publisher}{MIT Press},
  \bibinfo{address}{Cambridge, MA}, \bibinfo{year}{2005}).

\bibitem{Turing50}
\bibinfo{author}{Turing, A.~M.}
\newblock \bibinfo{title}{Computing machinery and intelligence}.
\newblock \emph{\bibinfo{journal}{Mind}} \textbf{\bibinfo{volume}{59}},
  \bibinfo{pages}{433--460} (\bibinfo{year}{1950}).

\bibitem{WengRepRev12}
\bibinfo{author}{Weng, J.}
\newblock \bibinfo{title}{Symbolic models and emergent models: A review}.
\newblock \emph{\bibinfo{journal}{{IEEE Trans. Autonomous Mental Development}}}
  \textbf{\bibinfo{volume}{4}}, \bibinfo{pages}{29--53} (\bibinfo{year}{2012}).

\bibitem{Russell10}
\bibinfo{author}{Russell, S.} \& \bibinfo{author}{Norvig, P.}
\newblock \emph{\bibinfo{title}{Artificial Intelligence: A Modern Approach}}
  (\bibinfo{publisher}{Prentice-Hall}, \bibinfo{address}{Upper Saddle River,
  New Jersey}, \bibinfo{year}{2010}), \bibinfo{edition}{3rd} edn.

\bibitem{Minsky91}
\bibinfo{author}{Minsky, M.}
\newblock \bibinfo{title}{Logical versus analogical or symbolic versus
  connectionist or neat versus scruffy}.
\newblock \emph{\bibinfo{journal}{{AI} Magazine}}
  \textbf{\bibinfo{volume}{12}}, \bibinfo{pages}{34--51}
  (\bibinfo{year}{1991}).

\bibitem{Lenat95b}
\bibinfo{author}{Lenat, D.~B.}, \bibinfo{author}{Miller, G.} \&
  \bibinfo{author}{Yokoi, T.~T.}
\newblock \bibinfo{title}{{CYC}, {WordNet}, and {EDR}: Critiques and
  responses}.
\newblock \emph{\bibinfo{journal}{Communications of the {ACM}}}
  \textbf{\bibinfo{volume}{38}}, \bibinfo{pages}{45--48}
  (\bibinfo{year}{1995}).

\bibitem{Gome14}
\bibinfo{author}{Gomes, L.}
\newblock \bibinfo{title}{Machine-learning maestro {Michael Jordan} on the
  delusions of big data and other huge engineering efforts}.
\newblock \emph{\bibinfo{journal}{IEEE Spectrum}}  (\bibinfo{year}{Online
  article posted Oct. 20, 2014}).

\bibitem{Rumelhart86}
\bibinfo{author}{Rumelhart, D.~E.}, \bibinfo{author}{McClelland, J.~L.} \&
  \bibinfo{author}{{the PDP Research Group}}.
\newblock \emph{\bibinfo{title}{Parallel Distributed Processing}},
  vol.~\bibinfo{volume}{1} (\bibinfo{publisher}{MIT Press},
  \bibinfo{address}{Cambridge, Massachusetts}, \bibinfo{year}{1986}).

\bibitem{McClelland86}
\bibinfo{editor}{McClelland, J.~L.}, \bibinfo{editor}{Rumelhart, D.~E.} \&
  \bibinfo{editor}{{The PDP Research Group}} (eds.)
  \emph{\bibinfo{title}{Parallel Distributed Processing}},
  vol.~\bibinfo{volume}{2} (\bibinfo{publisher}{MIT Press},
  \bibinfo{address}{Cambridge, Massachusetts}, \bibinfo{year}{1986}).

\bibitem{Frishna17}
\bibinfo{author}{Krishna, R.} \emph{et~al.}
\newblock \bibinfo{title}{Visual genome}.
\newblock \emph{\bibinfo{journal}{International Journal of Computer Vision}}
  \textbf{\bibinfo{volume}{123}}, \bibinfo{pages}{32--73}
  (\bibinfo{year}{2017}).

\bibitem{Funahashi89}
\bibinfo{author}{Funahashi, K.~I.}
\newblock \bibinfo{title}{On the approximate realization of continuous mappings
  by neural networks}.
\newblock \emph{\bibinfo{journal}{Neural Network}}
  \textbf{\bibinfo{volume}{2}}, \bibinfo{pages}{183--192}
  (\bibinfo{year}{1989}).

\bibitem{Poggio90a}
\bibinfo{author}{Poggio, T.} \& \bibinfo{author}{Girosi, F.}
\newblock \bibinfo{title}{Networks for approximation and learning}.
\newblock \emph{\bibinfo{journal}{Proceedings of The IEEE}}
  \textbf{\bibinfo{volume}{78}}, \bibinfo{pages}{1481--1497}
  (\bibinfo{year}{1990}).

\bibitem{Kohonen01}
\bibinfo{author}{Kohonen, T.}
\newblock \emph{\bibinfo{title}{Self-Organizing Maps}}
  (\bibinfo{publisher}{Springer-Verlag}, \bibinfo{address}{Berlin},
  \bibinfo{year}{2001}), \bibinfo{edition}{3rd} edn.

\bibitem{Fukushima80}
\bibinfo{author}{Fukushima, K.}
\newblock \bibinfo{title}{Neocognitron: {A} self-organizing neural network
  model for a mechanism of pattern recognition unaffected by shift in
  position}.
\newblock \emph{\bibinfo{journal}{Biological Cybernetics}}
  \textbf{\bibinfo{volume}{36}}, \bibinfo{pages}{193--202}
  (\bibinfo{year}{1980}).

\bibitem{Oja03}
\bibinfo{author}{Oja, M.}, \bibinfo{author}{Kaski, S.} \&
  \bibinfo{author}{Kohunen, T.}
\newblock \bibinfo{title}{Bibliography self-organizing maps (som) papers: 1998
  - 2001 addendum}.
\newblock \emph{\bibinfo{journal}{Neural Computing Surveys}}
  \textbf{\bibinfo{volume}{3}}, \bibinfo{pages}{1--156} (\bibinfo{year}{2003}).

\bibitem{WengCresIJCV97}
\bibinfo{author}{Weng, J.}, \bibinfo{author}{Ahuja, N.} \&
  \bibinfo{author}{Huang, T.~S.}
\newblock \bibinfo{title}{Learning recognition and segmentation using the
  {Cresceptron}}.
\newblock \emph{\bibinfo{journal}{{International Journal of Computer Vision}}}
  \textbf{\bibinfo{volume}{25}}, \bibinfo{pages}{109--143}
  (\bibinfo{year}{1997}).

\bibitem{Fukushima83}
\bibinfo{author}{Fukushima, K.}, \bibinfo{author}{Miyake, S.} \&
  \bibinfo{author}{Ito, T.}
\newblock \bibinfo{title}{Neocognitron: {A} neural network model for a
  mechanism of visual pattern recognition}.
\newblock \emph{\bibinfo{journal}{{IEEE Trans. Systems, Man and Cybernetics}}}
  \textbf{\bibinfo{volume}{13}}, \bibinfo{pages}{826--834}
  (\bibinfo{year}{1983}).

\bibitem{Serre06}
\bibinfo{author}{Serre, T.}, \bibinfo{author}{Poggio, T.},
  \bibinfo{author}{Riesenhuber, M.}, \bibinfo{author}{Wolf, L.} \&
  \bibinfo{author}{Bileschi, S.}
\newblock \bibinfo{title}{High-performance vision system exploiting key
  features of visual cortex}.
\newblock \emph{\bibinfo{journal}{US Patent}}
  \textbf{\bibinfo{volume}{US7606777B2}} (\bibinfo{year}{2006}).

\bibitem{Fei-Fei06}
\bibinfo{author}{Fei-Fei, L.}, \bibinfo{author}{Fergus, R.} \&
  \bibinfo{author}{Perona, P.}
\newblock \bibinfo{title}{One-shot learning of object categories}.
\newblock \emph{\bibinfo{journal}{{IEEE Trans. Pattern Analysis and Machine
  Intelligence}}} \textbf{\bibinfo{volume}{28}}, \bibinfo{pages}{594--611}
  (\bibinfo{year}{2006}).

\bibitem{WengAMDNL-2-2012}
\bibinfo{author}{Weng, J.}
\newblock \bibinfo{title}{Dialog initiation: Modeling amd: Closed skull or
  not?}
\newblock \emph{\bibinfo{journal}{IEEE CIS Autonomous Mental Development
  Newsletter}} \textbf{\bibinfo{volume}{9}}, \bibinfo{pages}{10--11}
  (\bibinfo{year}{2012}).

\bibitem{Werbos94}
\bibinfo{author}{Werbos, P.~J.}
\newblock \emph{\bibinfo{title}{The Roots of Backpropagation: From Ordered
  Derivatives to Neural Networks and Political Forecasting}}
  (\bibinfo{publisher}{Wiley}, \bibinfo{address}{Chichester},
  \bibinfo{year}{1994}).

\bibitem{LeCun98}
\bibinfo{author}{LeCun, Y.}, \bibinfo{author}{Bottou, L.},
  \bibinfo{author}{Bengio, Y.} \& \bibinfo{author}{Haffner, P.}
\newblock \bibinfo{title}{Gradient-based learning applied to document
  recognition}.
\newblock \emph{\bibinfo{journal}{Proceedings of IEEE}}
  \textbf{\bibinfo{volume}{86}}, \bibinfo{pages}{2278--2324}
  (\bibinfo{year}{1998}).

\bibitem{Krizhevsky12}
\bibinfo{author}{Krizhevsky, A.}, \bibinfo{author}{Sutskever, I.} \&
  \bibinfo{author}{Hinton, G.~E.}
\newblock \bibinfo{title}{Imagenet classification with deep convolutional
  neural networks}.
\newblock In \emph{\bibinfo{booktitle}{Advances in Neural Information
  Processing Systems 25}}, \bibinfo{pages}{1106--1114} (\bibinfo{year}{2012}).

\bibitem{LeCun15}
\bibinfo{author}{LeCun, Y.}, \bibinfo{author}{Bengio, L.} \&
  \bibinfo{author}{Hinton, G.}
\newblock \bibinfo{title}{Deep learning}.
\newblock \emph{\bibinfo{journal}{Nature}} \textbf{\bibinfo{volume}{521}},
  \bibinfo{pages}{436--444} (\bibinfo{year}{2015}).

\bibitem{Mnih15}
\bibinfo{author}{Mnih, V.} \emph{et~al.}
\newblock \bibinfo{title}{Human-level control through deep reinforcement
  learning}.
\newblock \emph{\bibinfo{journal}{Nature}} \textbf{\bibinfo{volume}{518}},
  \bibinfo{pages}{529--533} (\bibinfo{year}{2015}).

\bibitem{Silver16}
\bibinfo{author}{Silver, D.} \emph{et~al.}
\newblock \bibinfo{title}{Mastering the game of go with deep neural networks
  and tree search}.
\newblock \emph{\bibinfo{journal}{Nature}} \textbf{\bibinfo{volume}{529}},
  \bibinfo{pages}{484--489} (\bibinfo{year}{2016}).

\bibitem{Graves16}
\bibinfo{author}{Graves, A.} \emph{et~al.}
\newblock \bibinfo{title}{Hybrid computing using a neural network with dynamic
  external memory}.
\newblock \emph{\bibinfo{journal}{Nature}} \textbf{\bibinfo{volume}{538}},
  \bibinfo{pages}{471--476} (\bibinfo{year}{2016}).

\bibitem{Silver17}
\bibinfo{author}{Silver, D.} \emph{et~al.}
\newblock \bibinfo{title}{Mastering the game of go without human knowledge}.
\newblock \emph{\bibinfo{journal}{Nature}} \bibinfo{pages}{354--359}
  (\bibinfo{year}{2017}).

\bibitem{McKinney20}
\bibinfo{author}{McKinney, S.~M.} \emph{et~al.}
\newblock \bibinfo{title}{International evaluation of an {AI} system for breast
  cancer screening}.
\newblock \emph{\bibinfo{journal}{Nature}} \textbf{\bibinfo{volume}{577}},
  \bibinfo{pages}{89--94} (\bibinfo{year}{2020}).

\bibitem{Senior20}
\bibinfo{author}{Senior, A.~W.} \emph{et~al.}
\newblock \bibinfo{title}{Improved protein structure prediction using
  potentials from deep learning}.
\newblock \emph{\bibinfo{journal}{Nature}} \textbf{\bibinfo{volume}{577}},
  \bibinfo{pages}{706--710} (\bibinfo{year}{2020}).

\bibitem{Bellemare20}
\bibinfo{author}{Bellemare, M.~G.} \emph{et~al.}
\newblock \bibinfo{title}{Autonomous navigation of stratospheric balloons using
  reinforcement learning}.
\newblock \emph{\bibinfo{journal}{Nature}} \textbf{\bibinfo{volume}{588}},
  \bibinfo{pages}{77--82} (\bibinfo{year}{2020}).

\bibitem{Ecoffet21}
\bibinfo{author}{Ecoffet, A.}, \bibinfo{author}{Huizinga, J.},
  \bibinfo{author}{Lehman, J.}, \bibinfo{author}{Stanley, K.~O.} \&
  \bibinfo{author}{Clune, J.}
\newblock \bibinfo{title}{First return, then explore}.
\newblock \emph{\bibinfo{journal}{Nature}} \textbf{\bibinfo{volume}{590}},
  \bibinfo{pages}{580--586} (\bibinfo{year}{2021}).

\bibitem{Saggio21}
\bibinfo{author}{Saggio, V.} \emph{et~al.}
\newblock \bibinfo{title}{Experimental quantum speed-up in reinforcement
  learning agents}.
\newblock \emph{\bibinfo{journal}{Nature}} \textbf{\bibinfo{volume}{591}},
  \bibinfo{pages}{229--233} (\bibinfo{year}{2021}).

\bibitem{WillettText21}
\bibinfo{author}{Willett, F.~R.}, \bibinfo{author}{Avansino, D.~T.},
  \bibinfo{author}{Hochberg, L.~R.}, \bibinfo{author}{Henderson, J.~M.} \&
  \bibinfo{author}{Shenoy, K.~V.}
\newblock \bibinfo{title}{High-performance brain-to-text communication via
  handwriting}.
\newblock \emph{\bibinfo{journal}{Nature}} \textbf{\bibinfo{volume}{593}},
  \bibinfo{pages}{249--254} (\bibinfo{year}{2021}).

\bibitem{Slonim21}
\bibinfo{author}{Slonim, N.} \emph{et~al.}
\newblock \bibinfo{title}{An autonomous debating system}.
\newblock \emph{\bibinfo{journal}{Nature}} \textbf{\bibinfo{volume}{591}},
  \bibinfo{pages}{379--384} (\bibinfo{year}{2021}).

\bibitem{Mirhoseini21}
\bibinfo{author}{Mirhoseini, A.} \emph{et~al.}
\newblock \bibinfo{title}{A graph placement methodology for fast chip design}.
\newblock \emph{\bibinfo{journal}{Nature}} \textbf{\bibinfo{volume}{594}},
  \bibinfo{pages}{207--212} (\bibinfo{year}{2021}).

\bibitem{Lu21}
\bibinfo{author}{Lu, M.~Y.} \emph{et~al.}
\newblock \bibinfo{title}{{AI}-based pathology predicts origins for cancers of
  unknown primary}.
\newblock \emph{\bibinfo{journal}{Nature}} \textbf{\bibinfo{volume}{594}},
  \bibinfo{pages}{106--110} (\bibinfo{year}{2021}).

\bibitem{Warnat21}
\bibinfo{author}{Warnat-Herresthal, S.} \emph{et~al.}
\newblock \bibinfo{title}{Swarm learning for decentralized and confidential
  clinical machine learning}.
\newblock \emph{\bibinfo{journal}{Nature}} \textbf{\bibinfo{volume}{594}},
  \bibinfo{pages}{265--270} (\bibinfo{year}{2021}).

\bibitem{WengScience}
\bibinfo{author}{Weng, J.} \emph{et~al.}
\newblock \bibinfo{title}{Autonomous mental development by robots and animals}.
\newblock \emph{\bibinfo{journal}{Science}} \textbf{\bibinfo{volume}{291}},
  \bibinfo{pages}{599--600} (\bibinfo{year}{2001}).

\bibitem{McClelland07}
\bibinfo{author}{Mcclelland, J.~L.}, \bibinfo{author}{Plunkett, K.} \&
  \bibinfo{author}{Weng, J.}
\newblock \bibinfo{title}{Guest editorial: Convergent approaches to the
  understanding of autonomous mental development}.
\newblock \emph{\bibinfo{journal}{IEEE Trans. on Evolutionary Computation}}
  \textbf{\bibinfo{volume}{11}}, \bibinfo{pages}{133--136}
  (\bibinfo{year}{2007}).

\bibitem{WengIJIS15}
\bibinfo{author}{Weng, J.}
\newblock \bibinfo{title}{Brain as an emergent finite automaton: A theory and
  three theorems}.
\newblock \emph{\bibinfo{journal}{International Journal of Intelligent
  Science}} \textbf{\bibinfo{volume}{5}}, \bibinfo{pages}{112--131}
  (\bibinfo{year}{2015}).

\bibitem{WengIJCNN92}
\bibinfo{author}{Weng, J.}, \bibinfo{author}{Ahuja, N.} \&
  \bibinfo{author}{Huang, T.~S.}
\newblock \bibinfo{title}{Cresceptron: a self-organizing neural network which
  grows adaptively}.
\newblock In \emph{\bibinfo{booktitle}{Proc. Int'l Joint Conference on Neural
  Networks}}, vol.~\bibinfo{volume}{1}, \bibinfo{pages}{576--581}
  (\bibinfo{address}{Baltimore, Maryland}, \bibinfo{year}{1992}).

\bibitem{WengICCV93}
\bibinfo{author}{Weng, J.}, \bibinfo{author}{Ahuja, N.} \&
  \bibinfo{author}{Huang, T.~S.}
\newblock \bibinfo{title}{Learning recognition and segmentation of {3-D}
  objects from {2-D} images}.
\newblock In \emph{\bibinfo{booktitle}{Proc. IEEE 4th Int'l Conf. Computer
  Vision}}, \bibinfo{pages}{121--128} (\bibinfo{year}{1993}).

\bibitem{WengPlagiarismFaceBook2020}
\bibinfo{author}{Weng, J.}
\newblock \bibinfo{title}{Life is science (35): Did {Turing Awards} go to
  plagiarism?}
\newblock \bibinfo{howpublished}{Facebook blog} (\bibinfo{year}{2020}).
\newblock
  \bibinfo{note}{\url{www.facebook.com/juyang.weng/posts/10158305658699783}}.

\bibitem{WengPlagiarismYouTube2020}
\bibinfo{author}{Weng, J.}
\newblock \bibinfo{title}{Did {Turing Awards} go to plagiarism?}
\newblock \bibinfo{howpublished}{YouTube Video} (\bibinfo{year}{2020}).
\newblock \bibinfo{note}{1:05 hours, \url{https://youtu.be/EAhkH79TKFU}}.

\bibitem{WengWhy11}
\bibinfo{author}{Weng, J.}
\newblock \bibinfo{title}{Why have we passed {``neural networks do not abstract
  well''?}}
\newblock \emph{\bibinfo{journal}{Natural Intelligence: the INNS Magazine}}
  \textbf{\bibinfo{volume}{1}}, \bibinfo{pages}{13--22} (\bibinfo{year}{2011}).

\bibitem{Ji08}
\bibinfo{author}{Ji, Z.}, \bibinfo{author}{Weng, J.} \&
  \bibinfo{author}{Prokhorov, D.}
\newblock \bibinfo{title}{Where-what network 1: {``Where''} and {``What''}
  assist each other through top-down connections}.
\newblock In \emph{\bibinfo{booktitle}{Proc. IEEE Int'l Conference on
  Development and Learning}}, \bibinfo{pages}{61--66}
  (\bibinfo{address}{Monterey, CA}, \bibinfo{year}{2008}).

\bibitem{Guo15}
\bibinfo{author}{Guo, Q.}, \bibinfo{author}{Wu, X.} \& \bibinfo{author}{Weng,
  J.}
\newblock \bibinfo{title}{Cross-domain and within-domain synaptic maintenance
  for autonomous development of visual areas}.
\newblock In \emph{\bibinfo{booktitle}{Proc. the Fifth Joint IEEE International
  Conference on Development and Learning and on Epigenetic Robotics}},
  \bibinfo{pages}{+1--6} (\bibinfo{address}{Providence, RI},
  \bibinfo{year}{2015}).

\bibitem{FellemanVanEssen91}
\bibinfo{author}{Felleman, D.~J.} \& \bibinfo{author}{{Van Essen}, D.~C.}
\newblock \bibinfo{title}{Distributed hierarchical processing in the primate
  cerebral cortex}.
\newblock \emph{\bibinfo{journal}{Cerebral Cortex}}
  \textbf{\bibinfo{volume}{1}}, \bibinfo{pages}{1--47} (\bibinfo{year}{1991}).

\bibitem{Super76}
\bibinfo{author}{Super, C.~M.}
\newblock \bibinfo{title}{Environmental effects on motor development: {A} case
  of {Africa} infant precocity}.
\newblock \emph{\bibinfo{journal}{Developmental Medicine and Child Neurology}}
  \textbf{\bibinfo{volume}{18}}, \bibinfo{pages}{561--567}
  (\bibinfo{year}{1976}).

\bibitem{Thoroughman}
\bibinfo{author}{Thoroughman, K.~A.} \& \bibinfo{author}{Taylor, J.~A.}
\newblock \bibinfo{title}{Rapid reshaping of human motor generalization}.
\newblock \emph{\bibinfo{journal}{Jounal of Neuroscience}}
  \textbf{\bibinfo{volume}{25}}, \bibinfo{pages}{8948--8953}
  (\bibinfo{year}{2005}).

\bibitem{Rizzolatti87}
\bibinfo{author}{Rizzotti, G.}, \bibinfo{author}{Riggio, L.},
  \bibinfo{author}{Dascola, I.} \& \bibinfo{author}{Umilta, C.}
\newblock \bibinfo{title}{Reorienting attention across the horizontal and
  vertical meridians: evidence in favor of a premotor theory of attention}.
\newblock \emph{\bibinfo{journal}{Neuropsychologia}}
  \textbf{\bibinfo{volume}{25}}, \bibinfo{pages}{31--40}
  (\bibinfo{year}{1987}).

\bibitem{Moore03}
\bibinfo{author}{Moore, T.}, \bibinfo{author}{Armstrong, K.~M.} \&
  \bibinfo{author}{Fallah, M.}
\newblock \bibinfo{title}{Visuomotor origins of covert spatial attention}.
\newblock \emph{\bibinfo{journal}{Neuron}} \textbf{\bibinfo{volume}{40}},
  \bibinfo{pages}{671--683} (\bibinfo{year}{2003}).

\bibitem{Iverson10}
\bibinfo{author}{Iverson, J.~M.}
\newblock \bibinfo{title}{Developing language in a developing body: the
  relationship between motor development and language development}.
\newblock \emph{\bibinfo{journal}{Journal of child language}}
  \textbf{\bibinfo{volume}{37}}, \bibinfo{pages}{229--261}
  (\bibinfo{year}{2010}).

\bibitem{WengSpace12}
\bibinfo{author}{Weng, J.} \& \bibinfo{author}{Luciw, M.}
\newblock \bibinfo{title}{Brain-like emergent spatial processing}.
\newblock \emph{\bibinfo{journal}{{IEEE Trans. Autonomous Mental Development}}}
  \textbf{\bibinfo{volume}{4}}, \bibinfo{pages}{161--185}
  (\bibinfo{year}{2012}).

\bibitem{WengTime}
\bibinfo{author}{Weng, J.}, \bibinfo{author}{Luciw, M.} \&
  \bibinfo{author}{Zhang, Q.}
\newblock \bibinfo{title}{Brain-like temporal processing: Emergent open
  states}.
\newblock \emph{\bibinfo{journal}{{IEEE Trans. Autonomous Mental Development}}}
  \textbf{\bibinfo{volume}{5}}, \bibinfo{pages}{89 -- 116}
  (\bibinfo{year}{2013}).

\bibitem{WengIJHR2020}
\bibinfo{author}{Weng, J.}
\newblock \bibinfo{title}{Autonomous programming for general purposes: Theory}.
\newblock \emph{\bibinfo{journal}{International Journal of Huamnoid Robotics}}
  \textbf{\bibinfo{volume}{17}}, \bibinfo{pages}{1--36} (\bibinfo{year}{2020}).

\bibitem{WengCAI-ICDL20}
\bibinfo{author}{Weng, J.}
\newblock \bibinfo{title}{Conscious intelligence requires developmental
  autonomous programming for general purposes}.
\newblock In \emph{\bibinfo{booktitle}{Proc. IEEE International Conference on
  Development and Learning and Epigenetic Robotics}}, \bibinfo{pages}{1--7}
  (\bibinfo{address}{Valparaiso, Chile}, \bibinfo{year}{2020}).

\bibitem{WengPSUTS21}
\bibinfo{author}{Weng, J.}
\newblock \bibinfo{title}{On post selections using test sets {(PSUTS)} in
  {AI}}.
\newblock In \emph{\bibinfo{booktitle}{Proc. International Joint Conference on
  Neural Networks}}, \bibinfo{pages}{1--8} (\bibinfo{address}{Shengzhen,
  China}, \bibinfo{year}{2021}).

\bibitem{WengPSUTS-ICDL21}
\bibinfo{author}{Weng, J.}
\newblock \bibinfo{title}{A developmental method that computes optimal networks
  without post-selections}.
\newblock In \emph{\bibinfo{booktitle}{Proc. IEEE International Conference on
  Development and Learning}}, \bibinfo{pages}{1--6} (\bibinfo{address}{Beijing,
  China}, \bibinfo{year}{2021}).

\bibitem{Russakovsky15}
\bibinfo{author}{Russakovsky, O.} \emph{et~al.}
\newblock \bibinfo{title}{{ImageNet} large scale visual recognition challenge}.
\newblock \emph{\bibinfo{journal}{International Journal of Computer Vision}}
  \textbf{\bibinfo{volume}{115}}, \bibinfo{pages}{211--252}
  (\bibinfo{year}{2015}).

\bibitem{WengAAAIFS18}
\bibinfo{author}{Weng, J.} \emph{et~al.}
\newblock \bibinfo{title}{Emergent {Turing} machines and operating systems for
  brain-like auto-programming for general purposes}.
\newblock In \emph{\bibinfo{booktitle}{Proc. AAAI 2018 Fall Symposium:
  Gathering for Artificial Intelligence and Natural Systems}},
  \bibinfo{pages}{1--7} (\bibinfo{address}{Arlington, Virginia},
  \bibinfo{year}{2018}).

\bibitem{Ballard}
\bibinfo{author}{Ballard, D.~H.} \& \bibinfo{author}{Brown, C.~M.}
\newblock \emph{\bibinfo{title}{Computer Vision}}
  (\bibinfo{publisher}{Prentice-Hall}, \bibinfo{address}{New Jersey},
  \bibinfo{year}{1982}).

\bibitem{Shapiro01}
\bibinfo{author}{Shapiro, L.} \& \bibinfo{author}{Stockman, G.}
\newblock \emph{\bibinfo{title}{Computer Vision}}
  (\bibinfo{publisher}{Addison-Wesley}, \bibinfo{address}{New York},
  \bibinfo{year}{2001}).

\bibitem{WengNAI2ed2019}
\bibinfo{author}{Weng, J.}
\newblock \emph{\bibinfo{title}{Natural and Artificial Intelligence:
  Introduction to Computational Brain-Mind}} (\bibinfo{publisher}{BMI Press},
  \bibinfo{address}{Okemos, Michigan}, \bibinfo{year}{2019}),
  \bibinfo{edition}{second} edn.

\bibitem{Karpathy14}
\bibinfo{author}{Karpathy, A.} \emph{et~al.}
\newblock \bibinfo{title}{Large-scale video classification with convolutional
  neural networks}.
\newblock In \emph{\bibinfo{booktitle}{Proc. Computer Vision and Pattern
  Recognition}}, \bibinfo{pages}{+1--8} (\bibinfo{address}{Columbus, Ohio},
  \bibinfo{year}{2014}).

\bibitem{WengLCA09}
\bibinfo{author}{Weng, J.} \& \bibinfo{author}{Luciw, M.}
\newblock \bibinfo{title}{Dually optimal neuronal layers: Lobe component
  analysis}.
\newblock \emph{\bibinfo{journal}{{IEEE Trans. Autonomous Mental Development}}}
  \textbf{\bibinfo{volume}{1}}, \bibinfo{pages}{68--85} (\bibinfo{year}{2009}).

\bibitem{WengIEEE-IS2014}
\bibinfo{author}{Weng, J.} \& \bibinfo{author}{Luciw, M.~D.}
\newblock \bibinfo{title}{Brain-inspired concept networks: Learning concepts
  from cluttered scenes}.
\newblock \emph{\bibinfo{journal}{IEEE Intelligent Systems Magazine}}
  \textbf{\bibinfo{volume}{29}}, \bibinfo{pages}{14--22}
  (\bibinfo{year}{2014}).

\bibitem{Wood76}
\bibinfo{author}{Wood, D.~J.}, \bibinfo{author}{Bruner, J.~S.} \&
  \bibinfo{author}{Ross, G.}
\newblock \bibinfo{title}{The role of tutoring in problem-solving}.
\newblock \emph{\bibinfo{journal}{Journal of Child Psychology and Psychiatry}}
  \bibinfo{pages}{89--100} (\bibinfo{year}{1976}).

\bibitem{BurrActiveLearning98}
\bibinfo{author}{Burr, S.}
\newblock \bibinfo{title}{Active learning literature survey}.
\newblock \emph{\bibinfo{journal}{Data Mining and Knowledge Discovery}}
  \textbf{\bibinfo{volume}{2}}, \bibinfo{pages}{121--167}
  (\bibinfo{year}{1998}).

\bibitem{Krizhevsky17}
\bibinfo{author}{Krizhevsky, A.}, \bibinfo{author}{Sutskever, I.} \&
  \bibinfo{author}{Hinton, G.~E.}
\newblock \bibinfo{title}{Imagenet classification with deep convolutional
  neural networks}.
\newblock \emph{\bibinfo{journal}{Communications of the {ACM}}}
  \textbf{\bibinfo{volume}{60}}, \bibinfo{pages}{84--90}
  (\bibinfo{year}{2017}).

\bibitem{JainDubes}
\bibinfo{author}{Jain, A.~K.} \& \bibinfo{author}{Dubes, R.~C.}
\newblock \emph{\bibinfo{title}{Algorithms for Clustering Data}}
  (\bibinfo{publisher}{Prentice-Hall}, \bibinfo{address}{New Jersey},
  \bibinfo{year}{1988}).

\bibitem{Wang11}
\bibinfo{author}{Wang, Y.}, \bibinfo{author}{Wu, X.} \& \bibinfo{author}{Weng,
  J.}
\newblock \bibinfo{title}{Synapse maintenance in the where-what network}.
\newblock In \emph{\bibinfo{booktitle}{Proc. Int'l Joint Conference on Neural
  Networks}}, \bibinfo{pages}{2823--2829} (\bibinfo{address}{San Jose, CA},
  \bibinfo{year}{2011}).

\bibitem{GuoIJCNN14}
\bibinfo{author}{Guo, Q.}, \bibinfo{author}{Wu, X.} \& \bibinfo{author}{Weng,
  J.}
\newblock \bibinfo{title}{{WWN-9}: Cross-domain synaptic maintenance and its
  application to object groups recognition}.
\newblock In \emph{\bibinfo{booktitle}{Proc. Int'l Joint Conference on Neural
  Networks}}, \bibinfo{pages}{+1--8} (\bibinfo{address}{Beijing, China},
  \bibinfo{year}{2014}).

\bibitem{GaoBEAN21}
\bibinfo{author}{Gao, Q.}, \bibinfo{author}{Ascoli, G.~A.} \&
  \bibinfo{author}{Zhao, L.}
\newblock \bibinfo{title}{{BEAN}: Interpretable and efficient learning with
  biologically-enhanced artificial neuronal assembly regularization}.
\newblock \emph{\bibinfo{journal}{Front. Neurorobot}} \bibinfo{pages}{1--13}
  (\bibinfo{year}{2021}).

\bibitem{Moravcik17}
\bibinfo{author}{Moravcik, M.} \emph{et~al.}
\newblock \bibinfo{title}{Deepstack: Expert-level artificial intelligence in
  heads-up no-limit poker}.
\newblock \emph{\bibinfo{journal}{Science}} \textbf{\bibinfo{volume}{356}},
  \bibinfo{pages}{508--513} (\bibinfo{year}{2017}).

\bibitem{WengFraudFaceBook2020}
\bibinfo{author}{Weng, J.}
\newblock \bibinfo{title}{Life is science (36): Did {Turing Awards} go to
  fraud?}
\newblock \bibinfo{howpublished}{Facebook blog} (\bibinfo{year}{2020}).
\newblock
  \bibinfo{note}{\url{www.facebook.com/juyang.weng/posts/10158319020739783}}.

\bibitem{WengFraudYouTube2020}
\bibinfo{author}{Weng, J.}
\newblock \bibinfo{title}{Did {Turing Awards} go to fraud?}
\newblock \bibinfo{howpublished}{YouTube Video} (\bibinfo{year}{2020}).
\newblock \bibinfo{note}{1:04 hours, \url{https://youtu.be/Rz6CFlKrx2k}}.

\bibitem{Serre07}
\bibinfo{author}{Serre, T.}, \bibinfo{author}{Wolf, L.},
  \bibinfo{author}{Bileschi, S.}, \bibinfo{author}{Riesenhuber, M.} \&
  \bibinfo{author}{Poggio, T.}
\newblock \bibinfo{title}{Robust object recognition with cortex-like
  mechanisms}.
\newblock \emph{\bibinfo{journal}{{IEEE Trans. Pattern Analysis and Machine
  Intelligence}}} \textbf{\bibinfo{volume}{29}}, \bibinfo{pages}{411--426}
  (\bibinfo{year}{2007}).

\bibitem{Schaul13}
\bibinfo{author}{Sermanet, P.}, \bibinfo{author}{Kavukcuoglu, K.},
  \bibinfo{author}{Chintala, S.} \& \bibinfo{author}{LeCun, Y.}
\newblock \bibinfo{title}{No more pesky learning rates}.
\newblock In \emph{\bibinfo{booktitle}{Proc. International Conference on
  Machine learning}}, \bibinfo{pages}{343--351} (\bibinfo{address}{Atlanta,
  GA}, \bibinfo{year}{2013}).

\bibitem{Dauphin14}
\bibinfo{author}{Dauphin, Y.~N.} \emph{et~al.}
\newblock \bibinfo{title}{Identifying and attacking the saddle point problem in
  high-dimensional non-convex optimization}.
\newblock In \emph{\bibinfo{booktitle}{Advances in Neural Information
  Processing Systems}}, \bibinfo{pages}{2933--2941} (\bibinfo{publisher}{Curran
  Associates, Inc.}, \bibinfo{address}{Montreal, Canada},
  \bibinfo{year}{2014}).

\bibitem{Srivastava14}
\bibinfo{author}{Srivastava, N.}, \bibinfo{author}{Hinton, G.~E.},
  \bibinfo{author}{Krizhevsky, K.}, \bibinfo{author}{Sutskever, I.} \&
  \bibinfo{author}{Salakhutdinov, R.}
\newblock \bibinfo{title}{Dropout: A simple way to prevent neural networks from
  overtting}.
\newblock \emph{\bibinfo{journal}{Journal of Machine Learning Research}}
  \textbf{\bibinfo{volume}{15}}, \bibinfo{pages}{1929--1958}
  (\bibinfo{year}{2014}).

\bibitem{Poggio20}
\bibinfo{author}{Poggio, T.}
\newblock \bibinfo{title}{Theoretical issues in deep networks}.
\newblock \emph{\bibinfo{journal}{Proceedings of thr National Academy of
  Sciences}} \textbf{\bibinfo{volume}{117}}, \bibinfo{pages}{30039--30045}
  (\bibinfo{year}{2020}).

\bibitem{WengIJCNN2020}
\bibinfo{author}{Weng, J.}, \bibinfo{author}{Zheng, Z.}, \bibinfo{author}{Wu,
  X.} \& \bibinfo{author}{Castro-Garcia, J.}
\newblock \bibinfo{title}{Auto-programming for general purposes: Theory and
  experiments}.
\newblock In \emph{\bibinfo{booktitle}{Proc. International Joint Conference on
  Neural Networks}}, \bibinfo{pages}{1--8} (\bibinfo{address}{Glasgow, UK},
  \bibinfo{year}{2020}).

\bibitem{Silver18}
\bibinfo{author}{Silver, D.} \emph{et~al.}
\newblock \bibinfo{title}{A general reinforcement learning algorithm that
  masters chess, shogi, and go through self-play}.
\newblock \emph{\bibinfo{journal}{Science}} \bibinfo{pages}{1140--1144}
  (\bibinfo{year}{2018}).

\bibitem{Schrittwieser20}
\bibinfo{author}{Schrittwieser, J.} \emph{et~al.}
\newblock \bibinfo{title}{Mastering atari, go, chess and shogi by planning with
  a learned model}.
\newblock \emph{\bibinfo{journal}{Science}} \textbf{\bibinfo{volume}{588}},
  \bibinfo{pages}{604--609} (\bibinfo{year}{2020}).

\bibitem{WengPatentDN-2}
\bibinfo{author}{Weng, J.}, \bibinfo{author}{Zheng, Z.} \& \bibinfo{author}{Wu,
  X.}
\newblock \bibinfo{title}{Developmental {Network Two}, its optimality, and
  emergent {Turing} machines}.
\newblock \bibinfo{howpublished}{U.S. Provisional Patent Application Serial
  Number: 62/624,898} (\bibinfo{year}{2018}).
\newblock \bibinfo{note}{Published}.

\bibitem{Knoll-ICDL21}
\bibinfo{author}{Knoll, J.~A.} \emph{et~al.}
\newblock \bibinfo{title}{Optimal developmental learning for multisensory and
  multi-teaching modalities}.
\newblock In \emph{\bibinfo{booktitle}{Proc. IEEE International Conference on
  Development and Learning}}, \bibinfo{pages}{1--6} (\bibinfo{address}{Beijing,
  China}, \bibinfo{year}{2021}).

\bibitem{WengJCS2020}
\bibinfo{author}{Weng, J.}
\newblock \bibinfo{title}{A unified hierarchy for {AI} and natural intelligence
  through auto-programming for general purposes}.
\newblock \emph{\bibinfo{journal}{Journal of Cognitive Science}}
  \textbf{\bibinfo{volume}{21}}, \bibinfo{pages}{53--102}
  (\bibinfo{year}{2020}).

\bibitem{Rabiner89a}
\bibinfo{author}{Rabiner, L.~R.}
\newblock \bibinfo{title}{A tutorial on hidden {Markov} models and selected
  applications in speech recognition}.
\newblock \emph{\bibinfo{journal}{Proceedings of IEEE}}
  \textbf{\bibinfo{volume}{77}}, \bibinfo{pages}{257--286}
  (\bibinfo{year}{1989}).

\bibitem{Zheng19}
\bibinfo{author}{Zheng, Z.}, \bibinfo{author}{Wu, X.} \& \bibinfo{author}{Weng,
  J.}
\newblock \bibinfo{title}{Emergent neural turing macine and its visual
  navigation}.
\newblock \emph{\bibinfo{journal}{Neural Networks}}
  \textbf{\bibinfo{volume}{110}}, \bibinfo{pages}{116--130}
  (\bibinfo{year}{2019}).

\bibitem{WuIJCNN20}
\bibinfo{author}{Wu, X.} \& \bibinfo{author}{Weng, J.}
\newblock \bibinfo{title}{Muscle vectors as temporally {``Dense Labels''}}.
\newblock In \emph{\bibinfo{booktitle}{Proc. International Joint Conference on
  Neural Networks}}, \bibinfo{pages}{1--8} (\bibinfo{address}{Glasgow, UK},
  \bibinfo{year}{2020}).

\bibitem{Carstro-GarciaIJCNN19}
\bibinfo{author}{Castro-Garcia, J.} \& \bibinfo{author}{Weng, J.}
\newblock \bibinfo{title}{Emergent multilingual language acquisition using
  developmental networks}.
\newblock In \emph{\bibinfo{booktitle}{Proc. International Joint Conf. Neural
  Networks}}, \bibinfo{pages}{+1--8} (\bibinfo{publisher}{IEEE Press},
  \bibinfo{address}{Budapest, Hungary}, \bibinfo{year}{2019}).

\bibitem{WuThink21}
\bibinfo{author}{Wu, X.} \& \bibinfo{author}{Weng, J.}
\newblock \bibinfo{title}{On machine thinking}.
\newblock In \emph{\bibinfo{booktitle}{Proc. International Joint Conf. Neural
  Networks}}, \bibinfo{pages}{1--8} (\bibinfo{publisher}{IEEE Press},
  \bibinfo{address}{Shenzhen, China}, \bibinfo{year}{2021}).

\bibitem{ZhengCVVT16}
\bibinfo{author}{Zheng, Z.} \& \bibinfo{author}{Weng, J.}
\newblock \bibinfo{title}{Mobile device based outdoor navigation with on-line
  learning neural network: a comparison with convolutional neural network}.
\newblock In \emph{\bibinfo{booktitle}{Proc. 7th Workshop on Computer Vision in
  Vehicle Technology (CVVT 2016) at CVPR 2016}}, \bibinfo{pages}{11--18}
  (\bibinfo{address}{Las Vega}, \bibinfo{year}{2016}).

\end{thebibliography}



\end{document}